%% file: main.tex
\documentclass{bmvc2k}

\input{not}

\hypersetup{
  pdftitle={When Neural Collapse Fails to Route: Routing Drift in Continual Learning},
  pdfauthor={Nicolas Larue}
}

\title{When Neural Collapse Fails to Route: Routing Drift in Continual Learning}

\addauthor{Nicolas Larue}{nicolas.larue@ensea.fr}{1}
\addinstitution{ETIS, CY Cergy Paris Universit\'e\\
\mbox{ENSEA}, CNRS, France}

\runninghead{N. Larue}{When Neural Collapse Fails to Route}

\begin{document}
\maketitle

\begin{abstract}
\input{sections/abstract}
\end{abstract}

\begin{center}
  \refstepcounter{figure}\label{fig:routing_logit_failure}
  \resizebox{0.74\linewidth}{!}{\input{figures/tikz/figure1_routing_logit_failure}}
  \parbox{0.84\linewidth}{\small
  Figure~\thefigure: \textbf{Routing drift as a logit-level failure.}
  The correct class can win within its task block while a different block contains the global argmax. Routing drift is therefore not a within-task class error; it is a cross-task score-comparability error.}
\end{center}
\clearpage

\input{sections/introduction}
\input{sections/related_work}
\input{sections/preliminaries}
\input{sections/problem}
\input{sections/method}
\input{sections/experiments}
\input{sections/discussion}
\input{sections/conclusion}

\clearpage
\begin{center}
  {\Large\bfseries Supplementary Material}
\end{center}
\setcounter{figure}{0}
\setcounter{table}{0}
\setcounter{equation}{0}
\renewcommand{\thefigure}{S\arabic{figure}}
\renewcommand{\thetable}{S\arabic{table}}
\renewcommand{\theequation}{S\arabic{equation}}
\input{sections/appendix}

\clearpage
\bibliography{refs}
\end{document}

%% file: not.tex
\usepackage{booktabs}
\usepackage{amsmath}
\usepackage{amssymb}
\usepackage{graphicx}
\usepackage{wrapfig}
\usepackage{placeins}
\usepackage{multirow}
\usepackage{colortbl}
\usepackage{xcolor}
\usepackage{xspace}
\usepackage{algorithm}
\usepackage{algpseudocode}
\usepackage{pifont}
\usepackage{tikz}
\usetikzlibrary{arrows.meta,calc,fit,positioning}
\usepackage[capitalise,nameinlink,noabbrev]{cleveref}
\hypersetup{hypertexnames=false}

\definecolor{trolleygrey}{RGB}{128,128,128}
\newcommand{\cmark}{\ding{51}}

\def\ie{\emph{i.e}\bmvaOneDot}

\newcommand{\probase}{ProNC\xspace}
\newcommand{\method}{RCR\xspace}
\newcommand{\methodfull}{Reliability-Calibrated Routing\xspace}
\newcommand{\methodhead}{\probase{} + \method}
\newcommand{\tabbest}[1]{\textbf{\boldmath #1}}
\newcommand{\tabsecond}[1]{\underline{#1}}
\newcommand{\tabnote}[1]{\par\footnotesize\emph{Note:} #1}
\newcommand{\oursrow}{\rowcolor{gray!12}}

\newcommand{\diaghead}[1]{\rotatebox[origin=c]{45}{\strut #1}}
\newcommand{\seg}[1]{\vspace{0.3mm}\noindent\textbf{#1}}

\newcommand{\notscore}[1]{s_{\mathrm{#1}}}

\newcommand{\scoreetf}{\notscore{ETF}}

\newcommand{\scoreldc}{\notscore{LDC}}
\newcommand{\scorerad}{\notscore{rad}}
\newcommand{\scorecomp}{\notscore{CompMix}}
\newcommand{\scorecompfix}{\notscore{CompMixFix}}
\newcommand{\scoreinvgrad}{\notscore{invgrad}}
\newcommand{\scorercr}{\notscore{RCR}}

\input{math_commands}

%% file: math_commands.tex
\newcommand{\C}{\mathcal{C}}
\newcommand{\T}{\mathcal{T}}
\newcommand{\D}{\mathcal{D}}

\newcommand{\inner}[2]{\left\langle #1,#2 \right\rangle}

\newcommand{\argmax}{\operatorname*{arg\,max}}

%% file: sections/abstract.tex
Progressive neural-collapse classifiers can learn strong task-local decision boundaries while failing the global Class-IL decision. We identify the failure as \emph{routing drift}: under an expanding equiangular tight frame (ETF) classifier, a held-out sample can remain separable within its true task block, yet receive its largest global score from another block. This produces high Task-IL accuracy but low Class-IL accuracy, and small replay buffers can mask the problem because stored samples are optimized repeatedly and often remain well routed. We formalize route accuracy directly from the global class argmax and use it to separate within-task discrimination from cross-task score comparability. This decomposition suggests a classifier-level correction: preserve the global ETF frame, but add class-local reliability evidence when transported class measures deviate from the neural-collapse ideal. We develop \methodfull (\method), a lightweight post-hoc classifier that composes standardized ETF alignment with radial plausibility around transported, NC-shrunk class representatives. Across three datasets and two fixed replay-buffer budgets, this NC-consistent correction gives the best Class-IL FAA on five of six settings and reduces Class-IL forgetting on all six. The largest gains occur in harder routing regimes: up to $+6.75$ FAA over the reported reference, and up to $+8.03$ FAA and $+8.62$ final route accuracy in paired evaluations, while preserving Task-IL performance. These results position routing drift as a measurable failure mode of ETF-based continual learning and show that reliability-calibrated class evidence can repair part of the lost global routing without retraining the representation. Code and evaluation artifacts are available at \href{https://towzeur.github.io/RCR/}{this URL}.

%% file: figures/tikz/figure1_routing_logit_failure.tex
\begin{tikzpicture}[
  x=1cm,y=1cm,
  font=\small,
  taskbox/.style={draw=black!18, rounded corners=2.5pt, fill=black!2, minimum width=3.05cm, minimum height=1.65cm},
  truebox/.style={draw=blue!45!black, rounded corners=2.5pt, fill=blue!5, line width=0.7pt, minimum width=3.05cm, minimum height=1.65cm},
  bar/.style={draw=none, rounded corners=0.7pt},
  note/.style={font=\footnotesize, text=black!65},
]

\definecolor{figblue}{RGB}{31,119,180}
\definecolor{figgreen}{RGB}{23,138,85}
\definecolor{figpurple}{RGB}{122,81,149}
\definecolor{figorange}{RGB}{208,135,112}
\definecolor{figred}{RGB}{180,35,24}

\node[note] at (5.05,2.02) {Example feature $z(x)$ with true label $y=37$ in task block $3$};

\node[taskbox] (t2) at (1.55,0.62) {};
\node[truebox] (t3) at (5.05,0.62) {};
\node[taskbox] (t8) at (8.55,0.62) {};

\node[font=\bfseries] at (1.55,1.28) {Task 2};
\node[font=\bfseries, text=blue!45!black] at (5.05,1.28) {True task 3};
\node[font=\bfseries] at (8.55,1.28) {Task 8};

\newcommand{\logitbar}[4]{%
  \fill[#4, bar] (#1-0.18,-0.18) rectangle (#1+0.18,#3);
  \node[font=\scriptsize] at (#1,-0.43) {#2};
}

\logitbar{0.92}{20}{0.26}{black!28}
\logitbar{1.55}{21}{0.44}{black!28}
\logitbar{2.18}{22}{0.34}{black!28}

\logitbar{4.42}{34}{0.52}{figpurple}
\logitbar{5.05}{35}{0.34}{black!28}
\logitbar{5.68}{37}{0.86}{figblue}
\node[font=\scriptsize, text=figblue] at (5.68,1.02) {$y$};

\logitbar{7.92}{80}{0.39}{black!28}
\logitbar{8.55}{82}{1.06}{figred}
\logitbar{9.18}{83}{0.41}{black!28}
\node[align=center, font=\footnotesize, text=figgreen] at (5.05,-0.92) {within true task\\$\argmax_{c:\tau(c)=3}s_c=37$};

\node[align=center, font=\footnotesize, text=figred] at (8.72,-0.94) {global decision\\$\argmax_{c\in\mathcal C_t}s_c=82$};

\node[draw=figgreen!50!black, fill=figgreen!6, rounded corners=2pt, minimum width=2.55cm, minimum height=0.44cm] at (3.45,-1.58) {Task-IL correct};
\node[draw=figred!65!black, fill=figred!6, rounded corners=2pt, minimum width=2.55cm, minimum height=0.44cm] at (6.65,-1.58) {Class-IL wrong};

\end{tikzpicture}

%% file: sections/introduction.tex
\section{Introduction}

Class-incremental learning (Class-IL) asks a model to classify among all classes observed so far, without receiving the task identity at test time. Task-incremental learning (Task-IL) is easier: the correct task block is known, so the classifier only chooses among classes inside that block. The gap between Class-IL and Task-IL therefore measures more than forgetting; it reveals whether evidence remains comparable across task blocks~\cite{mccloskey1989catastrophic,rebuffi2017icarl,buzzega2020dark}. Prior analyses decompose Class-IL into selecting the correct task and selecting the correct class within that task~\cite{kim2022theoreticalcl,kim2023learnabilitycl,lin2024tpl}. We study the score-level form of this decomposition inside progressive ETF classifiers.

Neural-collapse-inspired continual learners use equiangular tight frame (ETF) classifiers to impose a structured global geometry on class prototypes~\cite{papyan2020prevalence}. \probase is a representative example: it expands the ETF as new classes arrive and learns features aligned to the evolving frame~\cite{wang2026proNC}. This geometry yields strong Task-IL performance, which means task-local class separation is often preserved. Yet our diagnostics show that many remaining Class-IL errors are not same-task confusions. The correct class can be identifiable once the true task block is known, while the global ETF score assigns the sample to another block.

\Cref{fig:routing_logit_failure} illustrates this score-level failure. The true class has the largest logit inside its task block, so the Task-IL decision is correct, but another block contains the global maximum. We call this \emph{routing drift}. In ETF continual-learning runs, Task-IL can remain high while route accuracy collapses, indicating that the model keeps task-local discrimination but loses global route comparability.

This paper studies routing drift as a geometric failure mode of neural-collapse continual learning. The point is not that ETF classifiers fail to separate classes locally; rather, their global scores become unreliable across task blocks. We therefore ask whether Class-IL can be improved by correcting cross-task competition while preserving the task-local structure already learned by the progressive ETF model. Our answer is \methodfull (\method), a reliability-compositional score that combines global ETF alignment with radial plausibility around transported class measures. \method targets the inference geometry of progressive ETF classifiers: it keeps the global neural-collapse frame, adds local class reliability, and does not use an external task predictor. It is a post-hoc classifier-level correction, not a new representation learner; this scope is deliberate because it isolates the routing failure from the underlying training dynamics. The method is designed for progressive ETF classifiers, and extending the same reliability principle beyond explicit ETF frames is left open.

Our contribution is threefold. \textbf{(i)} We define route accuracy as a diagnostic read directly from the global ETF prediction, specializing the known Class-IL decomposition to progressive neural-collapse classifiers. \textbf{(ii)} We show that replay buffers can mask routing drift, so replay-fitted task calibration is not a reliable substitute for held-out routing evaluation. \textbf{(iii)} We introduce \method, a reliability-calibrated classifier that preserves the neural-collapse frame, models local class plausibility, gives the best Class-IL FAA on five of six dataset-buffer settings, and reduces Class-IL forgetting on all six while preserving Task-IL accuracy.

%% file: sections/related_work.tex
\section{Related Work}
\label{sec:related_work}

\paragraph{Class-incremental continual learning.}
Class-incremental learning evaluates a model after each task over all classes seen so far, without providing the task identity at test time~\cite{masana2022class}. Early and still influential approaches regularize important parameters or logits~\cite{kirkpatrick2017overcoming,zenke2017continual,li2017learning}, or rehearse previous data through direct replay~\cite{riemer2019learning,rolnick2019experience}, exemplar classifiers~\cite{rebuffi2017icarl}, gradient-aware replay~\cite{lopez2017gradient,chaudhry2018efficient,aljundi2019gradient}, and regularized replay~\cite{buzzega2020dark,boschini2022class}. Other approaches use cosine classifiers, pooled-output distillation, feature boosting, prompts, contrastive objectives, coreset selection, provable memory construction, and stability-inducing weight perturbations~\cite{hou2019learning,douillard2020podnet,wang2022foster,smith2023coda,cha2021co2l,wen2024provable,dang2025memory,tong2025coreset,eskandarstar}. These methods address forgetting in the representation and in the classifier, but Class-IL also requires scores from different tasks to remain comparable. Prior analyses separate this requirement into two events: selecting the correct task block and selecting the correct class inside that block~\cite{kim2022theoreticalcl,kim2023learnabilitycl,lin2024tpl}. We use this decomposition diagnostically, but do not train a task predictor. Route accuracy is read directly from the global class argmax, and \method changes the class score itself rather than fitting a separate task-id module.

\paragraph{Neural collapse and ETF classifiers.}
Neural collapse describes a terminal training geometry in which within-class variability shrinks, class means align with classifier weights, and these directions form a simplex ETF~\cite{papyan2020prevalence}. Subsequent work has analyzed its dynamics, layer-peeled models, imbalanced regimes, and feature-classifier alignment beyond standard balanced classification~\cite{han2022neural,fang2021exploring,yang2023neural,xie2023neural}. This geometry has motivated fixed ETF classifiers for continual learning, where the classifier is prescribed rather than learned from scratch at every task~\cite{NCT1}. Neural-collapse-inspired contrastive continual learning further uses predefined hard prototypes and simplex-structure distillation to reduce representational overlap and stabilize learning across tasks~\cite{montmaur2024ncc}. Progressive neural-collapse methods avoid committing to a final class set by expanding the ETF as new classes arrive, then using nearest-ETF inference in the current frame~\cite{wang2026proNC}. This design gives a simple global geometry for all classes seen so far. Our work starts from this trained progressive ETF model and asks a different question: after several feature-frame updates, is the nearest-ETF score still calibrated for both old and new classes? The answer motivates a classifier-level correction rather than a new ETF construction.

\paragraph{Prototype drift and local reliability.}
Prototype and cosine classifiers are widely used when class evidence is expressed by normalized features, class means, or learned weight vectors~\cite{wang2018cosface,rebuffi2017icarl}. They are also sensitive to representation drift: an old class mean estimated in one feature frame may no longer be reliable after the backbone changes. Learnable Drift Compensation (LDC) estimates a previous-to-current feature projector and applies it to old class prototypes before nearest-class-mean inference~\cite{gomezvilla2024ldc}. This line of work shows that prototype positions should be interpreted in the current feature frame, not as fixed historical measurements. A complementary issue is metric reliability. FeCAM models class distributions with covariance-aware distances, showing that a single isotropic nearest-prototype rule can be too restrictive when local class geometry differs across classes~\cite{goswami2023fecam}. We use these observations in a constrained way: transported prototypes and competitive radii provide local reliability evidence, while the explicit ETF classifier remains the global neural-collapse frame. Thus \method is not transported NCM, not a replay-fitted calibration head, and not a replacement for ProNC's progressive ETF.

\paragraph{Routing terminology.}
In many-task learning, task routing can denote an architectural mechanism that selects task-specific paths~\cite{strezoski2019taskrouting}; recent mixture-of-experts continual-learning papers similarly study expert assignment drift when samples are reassigned to newly added experts~\cite{zhao2026dymoe,guo2026starmoe}. These works use an explicit router, expert assignment, or task-specific execution path. Our setting is different: there is one shared classifier, no test-time task identity, and no learned routing head. Routing drift denotes inter-block score incomparability inside the same global classifier. This distinction is important experimentally: improving route accuracy here means changing the class evidence used by the global argmax, not adding a task oracle in front of the classifier.

%% file: sections/preliminaries.tex
\section{Preliminaries}

\paragraph{Continual learning protocol.}
We consider a sequence of tasks $\T=\{1,\ldots,T\}$. Task $t$ provides data $\D_t=\{(x_i,y_i)\}$ whose labels are disjoint from previous tasks. Let $\C_t$ be the set of classes observed after task $t$, and let $\tau(c)$ map class $c$ to its task block. In Class-IL, the model predicts among all classes in $\C_t$ without receiving task identity. In Task-IL, prediction is restricted to the true task block. We report final average accuracy (FAA) and final forgetting (FF). If $a_i^t$ is the accuracy on task $i$ after training task $t$, then
\begin{equation}
\mathrm{FAA}=\frac{1}{T}\sum_{i=1}^{T} a_i^T,
\qquad
\mathrm{FF}=\frac{1}{T-1}\sum_{i=1}^{T-1}\max_{1\le u<T} a_i^u-a_i^T.
\end{equation}
Primary notation is summarized in the supplementary material
(\cref{tab:notation}).

\paragraph{Route accuracy.}
Following the decomposition discussed in \cref{sec:related_work}, we define a route diagnostic directly from the global class prediction. For a sample $(x,y)$ and a global prediction $\hat y$,
\begin{align}
\mathrm{Class\text{-}IL}(x) &= \mathbb{1}[\hat y=y],\\
R(x) &= \mathbb{1}[\tau(\hat y)=\tau(y)].
\end{align}
Route accuracy is the expectation of $R(x)$ over the test set. It is not an oracle available to the method, but a measurement of whether the class argmax falls inside the correct task block. High Task-IL but low route accuracy means that features remain separable within the correct block, but the global classifier assigns them to another block.

\paragraph{Neural collapse and ETF inference.}
Neural collapse describes the terminal regime in which centered class means and classifier weights align to a simplex equiangular tight frame (ETF)~\cite{papyan2020prevalence}. We recall the standard NC1--NC4 definitions in the supplementary material (\cref{app:neural_collapse}). A $K$-class simplex ETF is a set of unit vectors $\{u_c\}_{c=1}^{K}$ satisfying
\begin{equation}
\inner{u_i}{u_j}=-\frac{1}{K-1}\quad \text{for } i\neq j.
\end{equation}
Under the neural-collapse idealization, classification reduces to nearest-prototype inference in the ETF frame:
\begin{equation}
\hat y_{\mathrm{ETF}}(x)=\argmax_{c\in\C_t}\inner{z(x)}{u_c},
\qquad
z(x)=\frac{f_\theta(x)}{\|f_\theta(x)\|}.
\end{equation}

\paragraph{Progressive Neural Collapse.}
\probase progressively constructs the ETF target during continual learning rather than fixing a global ETF in advance~\cite{wang2026proNC}. After task 1, it extracts an ETF aligned with the learned first-task class means. Before each later task, it expands the ETF by adding vertices for new classes while keeping the old geometry close to the previous frame. For task $t$, the reference training objective combines supervised learning, alignment to the expanded ETF, and feature distillation from the previous model:
\begin{equation}
\mathcal{L}_{\mathrm{ProNC}}
=\mathcal{L}_{\mathrm{ce}}
+\lambda_1\mathcal{L}_{\mathrm{align}}(z,u_y)
+\lambda_2\mathcal{L}_{\mathrm{distill}}(z,z^{-}),
\label{eq:pronc_loss}
\end{equation}
where $z^{-}$ is the normalized feature from the previous model. At evaluation, the linear classifier is replaced by nearest-ETF inference:
\begin{equation}
\scoreetf(x,c)=\inner{z(x)}{u_c},
\qquad
\hat y=\argmax_{c\in\C_t}\scoreetf(x,c).
\label{eq:pronc_inference}
\end{equation}
This nearest-ETF rule is the reference classifier analyzed below.
Definitions of $\mathcal{L}_{\mathrm{align}}$ and
$\mathcal{L}_{\mathrm{distill}}$ are given in the supplementary material
(\cref{app:implementation_details}).

%% file: sections/problem.tex
\section{The Routing Drift Problem}
\label{sec:routing-drift}

\probase gives every seen class an ETF vertex and compares normalized features
to all vertices in one shared frame. This is the right geometry for global
Class-IL: all class vertices have equal norm and equal pairwise angle, so the
classifier layout itself does not favor a task block. That same symmetry makes
the failure interpretable. If a wrong task block wins under an ETF head, the
problem is not an uneven classifier layout. It is a mismatch between the
symmetric decision rule and the class feature distributions that occupy it.
Such a mismatch can emerge after several tasks because old classes are
represented through replay and distillation, whereas new classes are observed
densely in the current task.

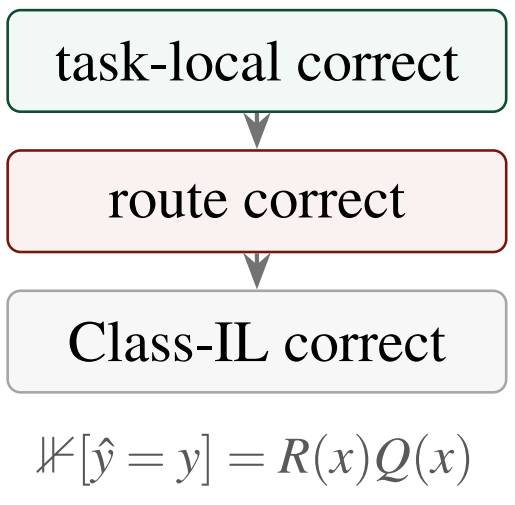
\begin{wrapfigure}{r}{0.31\linewidth}
  \centering
  \resizebox{0.265\textwidth}{!}{\input{figures/tikz/routing_problem_decomposition}}
  \caption{\textbf{Class-IL needs both events.}
  A prediction is globally correct only when the true task block wins and the
  correct class wins inside that block.}
  \label{fig:routing_problem_decomposition}
\end{wrapfigure}

To make the failure explicit, fix an evaluation stage and let $b$ index task
blocks. Write the best score inside block $b$ as
\begin{equation}
m_b(x)=\max_{c\in\C_t:\tau(c)=b}\scoreetf(x,c).
\end{equation}
As summarized in \cref{fig:routing_problem_decomposition}, Class-IL succeeds
only if the true block wins the cross-task competition and the correct class
wins inside that block. Task-IL measures only the second event. Hence a model
can have high Task-IL while failing globally whenever
\begin{equation}
\Delta_{\mathrm{route}}(x)
=m_{\tau(y)}(x)-\max_{b\neq \tau(y)}m_b(x)<0.
\label{eq:routing_failure}
\end{equation}
We call this \emph{routing drift}: features remain discriminative within their
own task block, but their global ETF scores no longer keep the correct block
ahead. In neural-collapse terms, this is a violation of equal class-measure
reliability, not a failure of the ETF geometry itself.

This gives a direct link between routing and Class-IL. Define the
block-restricted prediction
\begin{equation}
\hat y_b(x)=\argmax_{c\in\C_t:\tau(c)=b}\scoreetf(x,c).
\end{equation}
Let $R(x)=\mathbb{1}[\tau(\hat y)=\tau(y)]$ be the routing event, and let
$Q(x)=\mathbb{1}[\hat y_{\tau(y)}(x)=y]$ be the task-local event obtained by
restricting prediction to the true block. Exact Class-IL correctness requires
both events:
\begin{equation}
\mathbb{1}[\hat y=y] = R(x)\,Q(x).
\label{eq:class_route_factorization}
\end{equation}
Thus
\begin{equation}
\mathrm{Acc}_{\mathrm{Class\text{-}IL}}
=
\Pr(R=1)\,\Pr(Q=1\mid R=1).
\label{eq:class_route_expectation}
\end{equation}
Task-IL estimates the block-local event $Q$, whereas Class-IL also requires the
route $R$. When task-local discrimination is preserved, improving route accuracy
directly improves achievable Class-IL. This factorization is consistent with
prior analyses that decompose Class-IL into within-task prediction and
task-identity prediction~\cite{kim2022theoreticalcl,kim2023learnabilitycl,lin2024tpl}.
Here, route accuracy is not produced by a learned task-id predictor and is not
an oracle; it is read from the global ETF classifier's own argmax.

\begin{figure}[H]
  \centering
  \includegraphics[width=\linewidth]{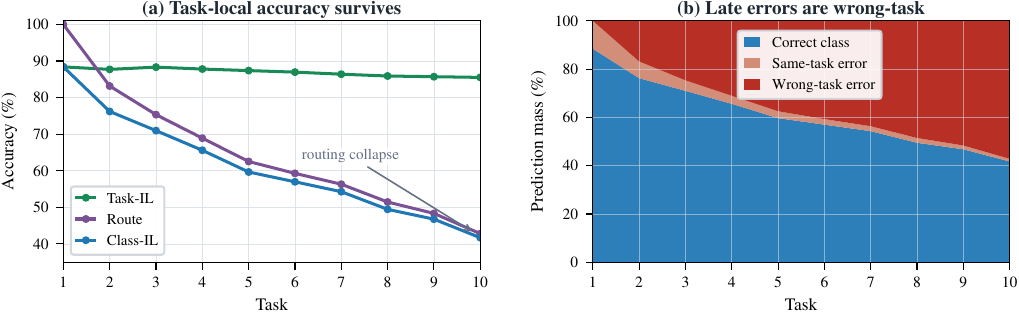}
  \caption{\textbf{Held-out errors are dominated by wrong-task routing.}
  On Seq-CIFAR-100 with buffer 200, Task-IL remains high across tasks, but
  Class-IL follows route accuracy downward. The right panel decomposes global
  predictions: late errors are mostly wrong-task routing rather than same-task
  class confusion.}
  \label{fig:routing_drift_pronc}
\end{figure}

An example block-score profile is shown in the supplementary material
(\cref{fig:appendix_task_block_score_profile}); the main diagnostic in
\cref{fig:routing_drift_pronc} shows that this is not an isolated sample-level
artifact but a sequence-level failure mode.

\begin{wrapfigure}[14]{r}{0.40\linewidth}
  \vspace{-0.3cm}
  \centering
  \includegraphics[width=0.38\textwidth]{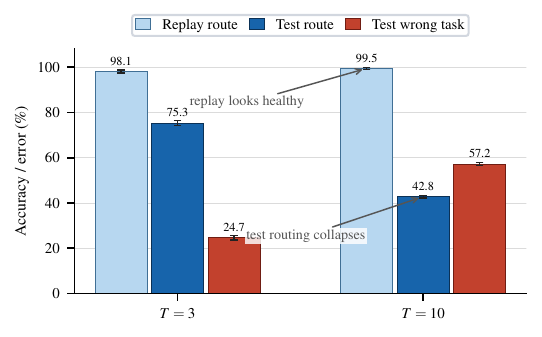}
  \caption{\textbf{Replay routing is optimistic.}
  Replay samples stay well routed while held-out test routing collapses on
  Seq-CIFAR-100/200; mean $\pm$ std over three seeds.}
  \label{fig:replay_blind_spot}
\end{wrapfigure}
Replay is also an unreliable route-calibration set. Stored samples are optimized
repeatedly and can remain close to their vertices: in
\cref{fig:replay_blind_spot}, replay routing is almost perfect at the final
task, while held-out test samples have only $42.81\%$ route accuracy and
$57.19\%$ wrong-task error. A useful correction should therefore preserve the
ETF frame, avoid fitting a task oracle from the buffer, and add local class
reliability directly to the global scoring rule. Replay labels reveal which
task a stored sample came from, but they do not measure how held-out old
samples compete against newer task blocks. The classifier must instead keep the
global ETF comparison intact and adjust class evidence only when the measured
class distribution has drifted away from its ideal NC geometry.
This is the motivation for \method, which we introduce in the next section.

%% file: figures/tikz/routing_problem_decomposition.tex
\begin{tikzpicture}[
  x=1cm,y=1cm,
  font=\small,
  note/.style={font=\footnotesize, text=black!65},
  arr/.style={-{Stealth[length=2mm]}, line width=0.65pt, draw=black!55},
]

\definecolor{figgreen}{RGB}{23,138,85}
\definecolor{figred}{RGB}{180,35,24}

\node[draw=figgreen!55!black, fill=figgreen!6, rounded corners=2pt,
      minimum width=2.7cm, minimum height=0.55cm] (local) at (1.55,1.78) {task-local correct};
\node[draw=figred!65!black, fill=figred!6, rounded corners=2pt,
      minimum width=2.7cm, minimum height=0.55cm] (route) at (1.55,1.02) {route correct};
\node[draw=black!35, fill=black!3, rounded corners=2pt,
      minimum width=2.7cm, minimum height=0.55cm] (classil) at (1.55,0.26) {Class-IL correct};
\draw[arr] (local) -- (route);
\draw[arr] (route) -- (classil);
\node[note, align=center] at (1.55,-0.38) {$\mathbb{1}[\hat y=y]=R(x)Q(x)$};

\end{tikzpicture}

%% file: sections/method.tex
\section{Method: Reliability-Calibrated Routing}

\subsection{From Routing Drift to Reliability-Calibrated Scores}

Routing drift calls for a classifier-level correction. The global ETF frame
should remain the shared comparison space, but each class score should also
reflect how reliable its local class measure is in the current feature frame.
Let $z=f_\theta(x)/\|f_\theta(x)\|$ and let $u_c$ be the ETF vertex for class
$c$. The reference score is
\begin{equation}
\scoreetf(x,c)=\inner{z}{u_c}.
\label{eq:etf_score}
\end{equation}
This score is globally comparable when all classes follow the same
neural-collapse geometry, but it treats every vertex as an equally reliable
point representative. \method keeps this global NC/ETF evidence and adds local
reliability around a transported prototype $p_c$ and radius $\rho_c$.

\subsection{Transported Class Measures}

\begin{figure}[t]
  \centering
  \subfigure[ETF route.]{
    \includegraphics[width=0.485\linewidth]{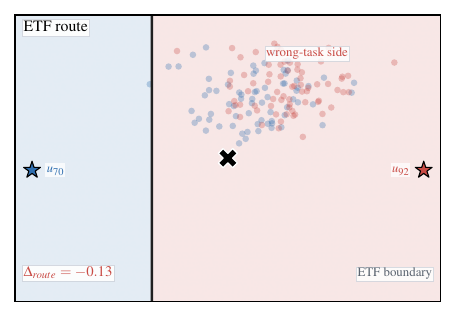}
  }\hfill
  \subfigure[\method route.]{
    \includegraphics[width=0.485\linewidth]{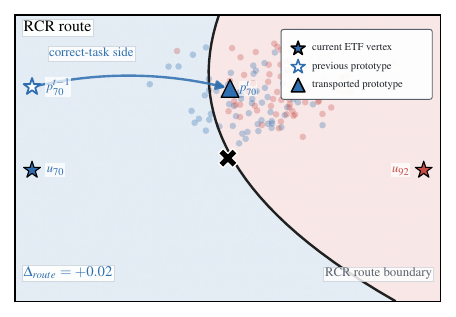}
  }
  \caption{\textbf{\method changes the route boundary without changing the
  backbone.}
  A held-out CIFAR-100 feature $z$ is shown in a local two-dimensional slice
  passing through the original high-dimensional feature, so $z$ has the same
  scores as in evaluation. The black curve is the route boundary
  between the correct task block and the ETF-predicted wrong block in that
  slice. (a) Under ETF scoring, $z$ lies in the red region: a wrong task block
  wins globally even though nearby features from the correct class remain
  coherent. (b) \method keeps ETF evidence but adds transported radial
  plausibility around $p_c^t$, moving the route boundary so the same
  feature falls in the correct-task region. Filled stars denote current ETF
  vertices, hollow stars denote previous prototypes, and triangles denote
  transported prototypes.
  Classes are selected only to visualize one corrected routing case; this
  selection is not used for training, hyperparameter choice, or evaluation.}
  \label{fig:radial_transport}
\end{figure}

Local reliability must be measured after feature drift. Following the
prototype-drift motivation of LDC~\cite{gomezvilla2024ldc}, we estimate a
previous-to-current linear transport. Unlike transported NCM, however, we use
transported prototypes only as local anchors; the ETF vertices remain the
global classifier. Thus $p_c$ is not a replacement for $u_c$. It is the
measured location of class $c$ in the current feature frame.
\Cref{fig:radial_transport} illustrates this separation.

At the end of task $t$, the previous frozen backbone and the current backbone
provide paired features $(z_i^{-},z_i)$ on the available training stream. We
fit the same form of ridge linear transport for all classes,
\begin{equation}
A_t=\operatorname*{arg\,min}_A
\sum_i \|z_i^{-}A-z_i\|_2^2+\lambda\|A\|_F^2.
\label{eq:prototype_transport}
\end{equation}
For old classes, the previous prototype is moved into the current frame by
\begin{equation}
p_c=\operatorname{norm}(p_c^{t-1}A_t).
\label{eq:transported_prototype}
\end{equation}
For new classes, $p_c$ is the normalized centroid of current-backbone features.
All class representatives now live in the same frame, but they are not used as
classifier weights.

\subsection{Radial Reliability}

The transported prototype gives a class location, not its reliability. Under
exact NC1, all classes are equally concentrated. In continual learning, replay
and drift make this concentration class-dependent. We measure this local NC1
violation while keeping $u_c$ as the global NC2/NC3 reference.
Let $d_{\cos}(a,b)=2-2\inner{a}{b}$ denote the cosine-distance convention used
by the radial head; for normalized vectors this is their squared Euclidean
distance, and the implementation only clamps it at zero for numerical
stability. Let $\mathcal B_c$ be the replay samples of class $c$ available
after task $t$. We estimate the raw class radius by
\begin{equation}
r_c=\frac{1}{|\mathcal B_c|}
\sum_{(x_i,y_i)\in\mathcal B_c}
d_{\cos}(z_i,p_c),
\qquad z_i=\frac{f_\theta(x_i)}{\|f_\theta(x_i)\|}.
\label{eq:raw_radius}
\end{equation}
If a class is absent from replay, we use the global mean radius as a fallback.

The raw radii cannot be unconstrained class scales. Since the ETF allocates
class directions uniformly, no class should win by claiming an arbitrarily large
region of the sphere. We therefore convert raw radii into a \emph{competitive}
radius:
\begin{equation}
\rho_c=\bar r\,
\frac{(r_c/\bar r)^{\alpha_\rho}}
{\frac{1}{|\C_t|}\sum_{j\in\C_t}(r_j/\bar r)^{\alpha_\rho}},
\qquad
\alpha_\rho=0.25,
\label{eq:competitive_radius}
\end{equation}
where $\bar r$ is the mean raw radius. This keeps the average radial budget
fixed while allowing measured dispersion to modulate local support; the exponent
$\alpha_\rho<1$ makes the modulation conservative. Let
$\bar\rho=\frac{1}{|\C_t|}\sum_{j\in\C_t}\rho_j$ denote the resulting mean
competitive radius. This normalization is not a softmax: increasing one class
radius increases its local support only relative to a fixed total radius budget,
so the other classes must compensate through the denominator.

Finally, using $p_c$ directly would let the local class measure replace the NC
reference. As a stabilization step, we \emph{retract} it toward $u_c$
according to relative reliability:
\begin{equation}
\tilde p_c=
\operatorname{norm}\!\left(
\alpha_c p_c+(1-\alpha_c)u_c
\right),
\qquad
\alpha_c=
\operatorname{clip}\!\left[
\left(\frac{\bar\rho}{\rho_c+\epsilon}\right)^\gamma,
\alpha_{\mathrm{lo}},\alpha_{\mathrm{hi}}
\right],
\label{eq:nc_shrink_proto}
\end{equation}
with $\gamma=0.5$ and $(\alpha_{\mathrm{lo}},\alpha_{\mathrm{hi}})=(0,0.95)$.
Classes with tight support keep more of their transported center, while
high-radius classes are pulled back toward the symmetric ETF vertex. The upper
clip prevents full trust in a drift-compensated prototype; the unconstrained
variant is reported in the supplement. The radial score is the negative
radius-normalized cosine distance to this NC-retracted center:
\begin{equation}
\scorerad(x,c)=
-\frac{d_{\cos}(z,\tilde p_c)}{\rho_c+\epsilon}.
\label{eq:radial_score}
\end{equation}
A class scores highly only when the query is close to its NC-retracted
prototype relative to its estimated radius. This term is class-local: it does
not predict a task block or introduce a separate router.

\subsection{Final Classifier}

ETF scores and radial scores have different units, so we standardize each
score vector per sample. This removes sample-specific scale and offset before
composition; the final decision depends on the relative class ordering supplied
by each evidence source, not on the arbitrary numerical units of cosine
alignment versus radius-normalized distance:
\begin{equation}
\mathcal{N}[s](x,c)=
\frac{s(x,c)-\frac{1}{|\C_t|}\sum_{j\in\C_t}s(x,j)}
{\operatorname{Std}_{j\in\C_t}\,s(x,j)+\epsilon}.
\label{eq:score_standardization}
\end{equation}
The final \method score is
\begin{equation}
\scorercr(x,c)=
(1-\omega)\,\mathcal{N}[\scoreetf](x,c)
+\omega\,\mathcal{N}[\scorerad](x,c),
\qquad
\omega=\frac{q}{q+1},
\label{eq:rcr_score}
\end{equation}
where $q$ is the number of classes per task. The scalar $\omega$ is fixed by
task granularity and shared by all classes and samples. It is not selected from
validation data: it increases the radial contribution when each task block
contains more competing classes, and keeps more ETF weight in small-block
settings such as Seq-CIFAR-10.

\method does not retrain the backbone, fit a task oracle, use test-time task
identity, or add a task-age prior. The only class-specific quantities are the
ETF vertex $u_c$, transported prototype $p_c$, and reliability radius $\rho_c$.
No scalar is selected per dataset: all reported datasets use the same fixed
rules $\alpha_\rho=0.25$, $\gamma=0.5$,
$(\alpha_{\mathrm{lo}},\alpha_{\mathrm{hi}})=(0,0.95)$, and
$\omega=q/(q+1)$.

\paragraph{NC consistency.}
\method is designed to be inactive at the exact neural-collapse fixed point.
Assume that transported class representatives coincide with ETF vertices,
$p_c=u_c$, and that all classes have the same radius, $\rho_c=\rho$. Then
$\tilde p_c=u_c$ by \cref{eq:nc_shrink_proto} for any clipped value of
$\alpha_c$, and
\begin{equation}
\scorerad(x,c)
=-\frac{2-2\inner{z}{u_c}}{\rho}
=\frac{2}{\rho}\scoreetf(x,c)-\frac{2}{\rho}.
\end{equation}
The radial score is therefore a positive affine transform of the ETF score.
Because \cref{eq:rcr_score} uses per-sample standardization,
$\mathcal N[\scorerad](x,\cdot)=
\mathcal N[\scoreetf](x,\cdot)$, and \method reduces exactly to the
normalized nearest-ETF classifier. Thus, \method changes decisions only when
progressive neural collapse is imperfect, \ie when transported class centers
deviate from ETF vertices or class reliabilities become heterogeneous.

%% file: sections/experiments.tex
\section{Experiments}

\subsection{Experimental Setup}

\seg{Benchmarks and evaluation setting.}
We evaluate on Seq-CIFAR-10, Seq-CIFAR-100~\cite{krizhevsky2009learning}, and
Seq-TinyImageNet~\cite{le2015tiny} with replay
buffers of 200 and 500 samples. Seq-CIFAR-10 is split into 5 tasks with 2
classes per task, while Seq-CIFAR-100 and Seq-TinyImageNet use 10 tasks with 10
and 20 classes per task, respectively. To isolate classifier geometry, all
paired head comparisons use the same learned progressive-ETF representations
and differ only in the evidence used for global prediction. Reproduced
artifacts follow the same ResNet-style backbone family as the reference
implementation~\cite{he2016deep}. Additional implementation details, including
the ProNC training objective and the values of $\lambda_1,\lambda_2$ used to
produce the checkpoints, are given in the supplementary material
(\cref{app:implementation_details}).

\seg{Baselines and paired evaluation.}
For the literature tables, all non-Ours baselines are taken from \cite{wang2026proNC}, which reports the same datasets, memory budgets, and
FAA/FF metrics. For the paired head comparison in
\cref{tab:validated_head_comparison}, we reproduced the reference model with
the official code over seeds $0,1,2$ and evaluated the ETF head and \methodhead
on the same artifacts. The reproduced ETF head is close to, but not identical
to, the published row: across the six dataset--buffer settings, the absolute
difference in final Class-IL FAA ranges from $0.29$ to $2.45$ points. We use
published rows for literature positioning and reproduced artifacts for paired
classifier comparisons.

\seg{Metrics.}
We report final average accuracy (FAA) and final forgetting (FF) in both
Class-IL and Task-IL. Class-IL evaluates the global classifier over all seen
classes. Task-IL restricts prediction to the ground-truth task block and
therefore measures task-local class discrimination. To quantify the failure mode
directly, we also report \emph{route accuracy}: the fraction of test samples for
which the globally predicted class belongs to the correct task block. A routing
correction should increase route accuracy and Class-IL while leaving Task-IL
nearly unchanged.

\subsection{Main Results}

\input{tables_tex/validated_head_comparison}

\subsubsection{Paired classifier comparison}

\Cref{tab:validated_head_comparison} compares the reference ETF score and
\methodhead on identical evaluation artifacts. \methodhead improves Class-IL FAA
and Route together, while Task-IL changes little. This is the signature
predicted by \cref{eq:class_route_factorization}: the gain comes from selecting
the correct task block, not from relearning within-task boundaries.

Across datasets and memory budgets, Class-IL gains track Route gains. After
progressive ETF training, the model often still separates classes within the
correct task block, but the global score assigns samples to competing blocks.
A compact visual summary and task-by-task diagnostic curves are provided in the
supplementary material
(\cref{fig:appendix_final_metric_grid,fig:appendix_all_diagnostic_curves}).

\subsubsection{Comparison against the state of the art}

The two tables below are for literature positioning. Their prior-method rows are
copied from the published reference table, while the Ours row is computed on our
three-seed reproduction. The perfectly paired comparison is therefore
\cref{tab:validated_head_comparison}, where both heads are evaluated on the
same reproduced checkpoints; the literature tables report whether the resulting
classifier correction reaches or exceeds the published numbers under the same
dataset and memory budgets.

\input{tables_tex/sota_200}
\input{tables_tex/sota_500}

\Cref{tab:sota_buffer_200,tab:sota_buffer_500} report the main comparison for
replay buffers of 200 and 500 samples. \methodhead sets the best Class-IL FAA
on five of six dataset-buffer settings and reduces Class-IL FF on all six
relative to the reported reference. These tables should be read as the effect
of a post-hoc, NC-consistent classifier correction applied to a trained
progressive ETF model, not as a replacement training algorithm for continual
learning. The largest gains appear on harder routing regimes, where many task
blocks compete globally.

On Seq-CIFAR-100, Class-IL FAA improves by $+6.75$ points with buffer 200 and
$+4.33$ with buffer 500; Class-IL FF is more than halved in both cases.
Seq-TinyImageNet improves under both budgets, and Seq-CIFAR-10/200 gains
$+4.78$ points. The only published-FAA exception is Seq-CIFAR-10/500, where the
reported reference is $0.27$ points higher. This is the benign regime predicted
by the mechanism: only five task blocks compete, Task-IL is near $97\%$, and
the Task-IL--Route gap is smallest. On matched checkpoints, \method still
improves this setting by $+1.52$ Class-IL FAA, $+1.62$ Route, and $-6.45$
Class-IL FF over the reproduced ETF head
(\cref{tab:validated_head_comparison}).

\subsection{Ablation}

We ablate the correction on Seq-CIFAR-100 with buffer 200, the hardest CIFAR
setting, while keeping the final checkpoint features fixed.

\input{tables_tex/component_ablation_compact}

\Cref{tab:component_ablation_compact} isolates the mechanism. Transported
radial evidence gives the largest gain over ETF-only scoring. A shared radius
removes class-dependent reliability and degrades Route. Removing NC retraction
causes a smaller but consistent drop. We therefore interpret the retraction as
an NC-consistent stabilization term rather than as the primary source of the
gain: the main improvement comes from transport, radial reliability, and the
competitive radius. Radial evidence without transport is inert in the composed
head: its center is the ETF vertex, so after score normalization it matches the
ETF decision.

We also test the simplest alternative explanation: a replay-fitted task bias
has direct access to replay labels, yet worsens held-out Route
(\cref{tab:replay_calibration_main}). This matches
\cref{fig:replay_blind_spot}: replay samples are too clean to expose test-time
routing drift.

\input{tables_tex/replay_calibration_main}

Full classifier-variant benchmarks and sensitivity studies are in the supplement.
We additionally include a small NCT diagnostic to check that the
route-correction signal is not tied to one progressive-ETF implementation. On
Seq-CIFAR-100 with buffer 200, an offline RCR-style ramp applied to a
reproduced NCT-style progressive ETF model improves Class-IL FAA by $+2.40$,
reduces Class-IL FF by $1.21$, and improves route accuracy by $+2.32$
(\cref{tab:nct_diagnostic}). This is a scope diagnostic, not a full
benchmark of NCT.

\subsection{Mechanistic Analysis}

The mechanism has two parts. The ETF score supplies a globally comparable NC
frame, while the transported radial score asks whether $z$ is locally plausible
under the current class measure. Score normalization makes the two evidences
commensurate before composition. In the exact NC limit, the radial score is a
positive affine transform of the ETF score, so \method changes no decisions.

\input{tables_tex/nc_cross_dataset_diag}

\Cref{tab:nc_cross_dataset_diag} makes the link quantitative across datasets:
the smallest gains occur when the ProNC task-local classifier already routes
well, and gains grow with the Task-IL--Route gap. Detailed stage-level NC
diagnostics, class-wise correlations, and true-route headroom are reported in
the supplementary material
(\cref{tab:nc_deviation_diag,tab:route_oracle_bounds}).

Seq-CIFAR-10 is closest to the benign regime: fewer task blocks compete,
Task-IL is near $97\%$, and route accuracy is already high. Accordingly,
\method gives smaller gains there, especially with buffer 500. Gains grow when
the Task-IL--Route gap is larger, supporting the intended interpretation:
\method is an NC-deviation correction, not a new within-task classifier. The
supplement further shows remaining true-route headroom and class-wise links
between center drift, radius growth, class age, and route error.

%% file: tables_tex/validated_head_comparison.tex
\begin{table}[h!]
  \centering
  \caption{\textbf{RCR improves the global decision by repairing route accuracy.}
  On the same frozen ProNC checkpoints, Class-IL and Route increase together while Task-IL remains nearly unchanged. Parentheses report deltas against $\scoreetf$; all entries are averaged over seeds $0,1,2$.}
  \label{tab:validated_head_comparison}
  \scriptsize
  \setlength{\tabcolsep}{3.2pt}
  \begin{tabular}{lllccccc}
    \toprule
    Dataset & Buffer & Score &
    \multicolumn{2}{c}{Class-IL} &
    \multicolumn{2}{c}{Task-IL} &
    Route \\
    \cmidrule(lr){4-5}\cmidrule(lr){6-7}
    & & & FAA $\uparrow$ & FF $\downarrow$ & FAA $\uparrow$ & FF $\downarrow$ & $\uparrow$ \\
    \midrule
    \multirow{4}{*}{CIFAR-10} & \multirow{2}{*}{200} & $\scoreetf$ & 66.14 & 31.47 & \tabbest{96.74} & \tabbest{0.62} & 66.58 \\
    & & $\scorercr$ (Ours) & \tabbest{70.36 {\scriptsize(+4.22)}} & \tabbest{22.85 {\scriptsize(-8.62)}} & 96.67 {\scriptsize(-0.07)} & 0.73 {\scriptsize(+0.11)} & \tabbest{70.82 {\scriptsize(+4.24)}} \\
    & \multirow{2}{*}{500} & $\scoreetf$ & 72.16 & 24.18 & \tabbest{96.82} & \tabbest{0.47} & 72.71 \\
    & & $\scorercr$ (Ours) & \tabbest{73.68 {\scriptsize(+1.52)}} & \tabbest{17.73 {\scriptsize(-6.45)}} & 96.81 {\scriptsize(-0.01)} & 0.54 {\scriptsize(+0.07)} & \tabbest{74.33 {\scriptsize(+1.62)}} \\
    \midrule
    \multirow{4}{*}{CIFAR-100} & \multirow{2}{*}{200} & $\scoreetf$ & 41.71 & 38.55 & 85.53 & 4.51 & 42.81 \\
    & & $\scorercr$ (Ours) & \tabbest{49.74 {\scriptsize(+8.03)}} & \tabbest{16.27 {\scriptsize(-22.28)}} & \tabbest{87.28 {\scriptsize(+1.75)}} & \tabbest{2.36 {\scriptsize(-2.15)}} & \tabbest{51.43 {\scriptsize(+8.62)}} \\
    & \multirow{2}{*}{500} & $\scoreetf$ & 48.65 & 23.16 & 85.54 & 4.29 & 50.14 \\
    & & $\scorercr$ (Ours) & \tabbest{53.27 {\scriptsize(+4.62)}} & \tabbest{10.18 {\scriptsize(-12.98)}} & \tabbest{87.74 {\scriptsize(+2.20)}} & \tabbest{2.02 {\scriptsize(-2.27)}} & \tabbest{55.04 {\scriptsize(+4.90)}} \\
    \midrule
    \multirow{4}{*}{TinyImageNet} & \multirow{2}{*}{200} & $\scoreetf$ & 25.77 & 43.45 & 68.30 & \tabbest{9.58} & 28.58 \\
    & & $\scorercr$ (Ours) & \tabbest{31.42 {\scriptsize(+5.65)}} & \tabbest{23.84 {\scriptsize(-19.61)}} & \tabbest{69.59 {\scriptsize(+1.29)}} & \tabbest{8.61 {\scriptsize(-0.97)}} & \tabbest{34.76 {\scriptsize(+6.18)}} \\
    & \multirow{2}{*}{500} & $\scoreetf$ & 26.61 & 41.16 & 67.71 & 9.78 & 29.37 \\
    & & $\scorercr$ (Ours) & \tabbest{32.24 {\scriptsize(+5.63)}} & \tabbest{22.47 {\scriptsize(-18.69)}} & \tabbest{70.16 {\scriptsize(+2.45)}} & \tabbest{7.90 {\scriptsize(-1.88)}} & \tabbest{35.46 {\scriptsize(+6.09)}} \\
    \bottomrule
  \end{tabular}
\end{table}

%% file: tables_tex/sota_200.tex
\begin{table*}[p]
\centering
\caption{Literature comparison for buffer size 200. FAA denotes final average accuracy, and FF denotes final forgetting.}
\label{tab:sota_buffer_200}
\scriptsize
\setlength{\tabcolsep}{3pt}
\resizebox{\textwidth}{!}{%
\begin{tabular}{lllcccccc}
\toprule
\multirow{2}{*}{Setting} & \multirow{2}{*}{Method} & \multirow{2}{*}{Venue} & \multicolumn{2}{c}{Seq-CIFAR-10} & \multicolumn{2}{c}{Seq-CIFAR-100} & \multicolumn{2}{c}{Seq-TinyImageNet} \\
\cmidrule(lr){4-5} \cmidrule(lr){6-7} \cmidrule(lr){8-9}
& & & FAA $\uparrow$ & FF $\downarrow$ & FAA $\uparrow$ & FF $\downarrow$ & FAA $\uparrow$ & FF $\downarrow$ \\
\midrule
\multirow{12}{*}{Class-IL} & ER~\citep{riemer2019learning} & ICLR 2019 & $44.79 \pm 1.86$ & $59.30 \pm 2.48$ & $21.78 \pm 0.48$ & $75.06 \pm 0.63$ & $8.49 \pm 0.16$ & $76.53 \pm 0.51$ \\
& iCaRL~\citep{rebuffi2017icarl} & CVPR 2017 & $49.02 \pm 3.20$ & $23.52 \pm 1.27$ & $28.00 \pm 0.91$ & $47.20 \pm 1.23$ & $7.53 \pm 0.79$ & \tabsecond{31.06 $\pm$ 1.91} \\
& DER~\citep{buzzega2020dark} & NeurIPS 2020 & $61.93 \pm 1.79$ & $35.79 \pm 2.59$ & $31.23 \pm 1.38$ & $62.72 \pm 2.69$ & $11.87 \pm 0.78$ & $64.83 \pm 1.48$ \\
& DER++~\citep{buzzega2020dark} & NeurIPS 2020 & $64.88 \pm 1.17$ & $32.59 \pm 2.32$ & $28.13 \pm 0.51$ & $60.99 \pm 1.52$ & $11.34 \pm 1.17$ & $73.47 \pm 1.23$ \\
& Co$^2$L~\citep{cha2021co2l} & ICCV 2021 & $51.27 \pm 1.86$ & $30.17 \pm 8.57$ & $18.09 \pm 0.49$ & $64.14 \pm 0.69$ & $12.95 \pm 0.06$ & $62.04 \pm 0.28$ \\
& CILA~\citep{wen2024provable} & ICML 2024 & $59.68 \pm 0.65$ & $37.52 \pm 5.84$ & $19.49 \pm 0.53$ & $64.01 \pm 0.18$ & $12.98 \pm 0.10$ & $63.11 \pm 0.61$ \\
& MNC\textsuperscript{3}L~\citep{dang2025memory} & WACV 2025 & $52.20 \pm 1.56$ & $33.74 \pm 1.65$ & $15.81 \pm 0.48$ & $62.51 \pm 0.40$ & $10.57 \pm 1.66$ & $59.68 \pm 2.02$ \\
& STAR~\citep{eskandarstar} & ICLR 2025 & $62.10 \pm 2.21$ & \tabbest{21.78 $\pm$ 3.16} & $18.29 \pm 2.58$ & $68.40 \pm 8.52$ & $11.55 \pm 3.23$ & $67.43 \pm 1.48$ \\
& CSReL~\citep{tong2025coreset} & ICLR 2025 & $37.46 \pm 1.57$ & $26.34 \pm 2.18$ & $29.06 \pm 1.00$ & $58.23 \pm 1.54$ & $18.14 \pm 3.10$ & $49.77 \pm 2.27$ \\
& NCT~\citep{NCT1} & ICLR 2023 & $51.59 \pm 0.41$ & \tabsecond{22.48 $\pm$ 19.50} & $26.38 \pm 0.57$ & \tabsecond{27.40 $\pm$ 1.65} & $10.95 \pm 1.45$ & $49.33 \pm 4.47$ \\
& ProNC~\citep{wang2026proNC} & ICLR 2026 & \tabsecond{65.58 $\pm$ 0.15} & $32.75 \pm 4.71$ & \tabsecond{42.99 $\pm$ 0.85} & $36.07 \pm 0.51$ & \tabsecond{27.44 $\pm$ 1.00} & $42.81 \pm 0.56$ \\
& \textbf{Ours} & This work & \tabbest{70.36 $\pm$ 0.62} & $22.85 \pm 1.45$ & \tabbest{49.74 $\pm$ 0.71} & \tabbest{16.27 $\pm$ 0.45} & \tabbest{31.42 $\pm$ 0.80} & \tabbest{23.84 $\pm$ 1.15} \\
\midrule
\multirow{12}{*}{Task-IL} & ER~\citep{riemer2019learning} & ICLR 2019 & $91.19 \pm 0.94$ & $6.07 \pm 1.09$ & $60.19 \pm 1.01$ & $27.38 \pm 1.46$ & $38.17 \pm 2.00$ & $40.47 \pm 1.54$ \\
& iCaRL~\citep{rebuffi2017icarl} & CVPR 2017 & $88.99 \pm 2.13$ & $25.34 \pm 1.64$ & $51.43 \pm 1.47$ & $36.20 \pm 1.85$ & $28.19 \pm 1.47$ & $42.47 \pm 2.47$ \\
& DER~\citep{buzzega2020dark} & NeurIPS 2020 & $91.40 \pm 0.92$ & $6.08 \pm 0.70$ & $63.09 \pm 1.09$ & $25.98 \pm 1.55$ & $40.22 \pm 0.67$ & $40.43 \pm 1.05$ \\
& DER++~\citep{buzzega2020dark} & NeurIPS 2020 & $91.92 \pm 0.60$ & $5.16 \pm 0.21$ & $66.80 \pm 0.41$ & $23.91 \pm 0.55$ & $43.06 \pm 1.16$ & $39.02 \pm 0.43$ \\
& Co$^2$L~\citep{cha2021co2l} & ICCV 2021 & $84.69 \pm 1.52$ & $2.91 \pm 5.25$ & $49.19 \pm 0.91$ & $36.81 \pm 0.63$ & $37.07 \pm 1.62$ & $40.75 \pm 3.55$ \\
& CILA~\citep{wen2024provable} & ICML 2024 & $91.36 \pm 0.08$ & $5.49 \pm 0.74$ & $53.93 \pm 1.02$ & $33.07 \pm 0.96$ & $37.32 \pm 1.87$ & $41.40 \pm 0.51$ \\
& MNC\textsuperscript{3}L~\citep{dang2025memory} & WACV 2025 & $85.94 \pm 0.22$ & $4.90 \pm 0.15$ & $43.91 \pm 0.76$ & $39.79 \pm 0.98$ & $32.78 \pm 3.54$ & $45.10 \pm 1.57$ \\
& STAR~\citep{eskandarstar} & ICLR 2025 & $93.54 \pm 1.48$ & $6.23 \pm 2.06$ & $58.53 \pm 13.06$ & $27.84 \pm 8.01$ & $41.70 \pm 0.95$ & $34.78 \pm 2.08$ \\
& CSReL~\citep{tong2025coreset} & ICLR 2025 & $69.22 \pm 3.03$ & $17.16 \pm 2.91$ & $66.99 \pm 0.35$ & $23.20 \pm 1.74$ & $45.04 \pm 5.86$ & $34.12 \pm 2.18$ \\
& NCT~\citep{NCT1} & ICLR 2023 & $80.63 \pm 0.46$ & $1.41 \pm 1.09$ & $75.75 \pm 0.17$ & $4.79 \pm 0.07$ & $52.71 \pm 4.12$ & $15.88 \pm 0.95$ \\
& ProNC~\citep{wang2026proNC} & ICLR 2026 & \tabbest{96.86 $\pm$ 0.10} & \tabbest{0.65 $\pm$ 0.08} & \tabsecond{85.63 $\pm$ 0.73} & \tabsecond{4.40 $\pm$ 0.82} & \tabsecond{69.09 $\pm$ 0.65} & \tabsecond{9.42 $\pm$ 0.58} \\
& \textbf{Ours} & This work & \tabsecond{96.67 $\pm$ 0.09} & \tabsecond{0.73 $\pm$ 0.19} & \tabbest{87.28 $\pm$ 0.25} & \tabbest{2.36 $\pm$ 0.14} & \tabbest{69.59 $\pm$ 0.99} & \tabbest{8.61 $\pm$ 1.09} \\
\bottomrule
\end{tabular}%
}
\end{table*}

%% file: tables_tex/sota_500.tex
\begin{table*}[p]
\centering
\caption{Literature comparison for buffer size 500. FAA denotes final average accuracy, and FF denotes final forgetting.}
\label{tab:sota_buffer_500}
\scriptsize
\setlength{\tabcolsep}{3pt}
\resizebox{\textwidth}{!}{%
\begin{tabular}{lllcccccc}
\toprule
\multirow{2}{*}{Setting} & \multirow{2}{*}{Method} & \multirow{2}{*}{Venue} & \multicolumn{2}{c}{Seq-CIFAR-10} & \multicolumn{2}{c}{Seq-CIFAR-100} & \multicolumn{2}{c}{Seq-TinyImageNet} \\
\cmidrule(lr){4-5} \cmidrule(lr){6-7} \cmidrule(lr){8-9}
& & & FAA $\uparrow$ & FF $\downarrow$ & FAA $\uparrow$ & FF $\downarrow$ & FAA $\uparrow$ & FF $\downarrow$ \\
\midrule
\multirow{12}{*}{Class-IL} & ER~\citep{riemer2019learning} & ICLR 2019 & $57.74 \pm 2.48$ & $43.22 \pm 2.10$ & $22.35 \pm 0.61$ & $73.08 \pm 0.78$ & $9.99 \pm 0.29$ & $75.21 \pm 0.54$ \\
& iCaRL~\citep{rebuffi2017icarl} & CVPR 2017 & $47.55 \pm 3.95$ & $28.20 \pm 2.41$ & $33.25 \pm 1.25$ & $40.99 \pm 1.02$ & $9.38 \pm 1.53$ & \tabsecond{37.30 $\pm$ 1.42} \\
& DER~\citep{buzzega2020dark} & NeurIPS 2020 & $70.51 \pm 1.67$ & $24.02 \pm 1.63$ & $41.36 \pm 1.76$ & $49.07 \pm 2.54$ & $17.75 \pm 1.14$ & $59.95 \pm 2.31$ \\
& DER++~\citep{buzzega2020dark} & NeurIPS 2020 & $72.70 \pm 1.36$ & $22.38 \pm 4.41$ & $38.20 \pm 1.00$ & $49.18 \pm 2.19$ & $19.38 \pm 1.41$ & $58.75 \pm 1.93$ \\
& Co$^2$L~\citep{cha2021co2l} & ICCV 2021 & $61.78 \pm 4.22$ & $17.79 \pm 3.36$ & $26.64 \pm 1.42$ & $48.60 \pm 1.34$ & $18.71 \pm 0.84$ & $49.64 \pm 1.14$ \\
& CILA~\citep{wen2024provable} & ICML 2024 & $67.82 \pm 0.33$ & $18.22 \pm 1.32$ & $31.27 \pm 0.17$ & $45.67 \pm 1.02$ & $18.09 \pm 0.49$ & $64.14 \pm 0.69$ \\
& MNC\textsuperscript{3}L~\citep{dang2025memory} & WACV 2025 & $52.20 \pm 1.56$ & $27.88 \pm 2.75$ & $22.29 \pm 0.18$ & $46.09 \pm 0.58$ & $11.52 \pm 0.01$ & $44.96 \pm 0.09$ \\
& STAR~\citep{eskandarstar} & ICLR 2025 & $69.15 \pm 3.53$ & $18.59 \pm 2.91$ & $28.45 \pm 1.70$ & $53.11 \pm 1.57$ & $15.19 \pm 2.61$ & $63.42 \pm 3.19$ \\
& CSReL~\citep{tong2025coreset} & ICLR 2025 & $40.82 \pm 4.09$ & $29.18 \pm 6.91$ & $39.22 \pm 1.70$ & $47.03 \pm 2.57$ & $22.13 \pm 0.35$ & $50.01 \pm 1.87$ \\
& NCT~\citep{NCT1} & ICLR 2023 & $60.93 \pm 0.94$ & \tabbest{13.82 $\pm$ 3.64} & $33.84 \pm 0.38$ & $24.37 \pm 5.42$ & $18.24 \pm 0.62$ & $50.90 \pm 1.16$ \\
& ProNC~\citep{wang2026proNC} & ICLR 2026 & \tabbest{73.95 $\pm$ 0.68} & $26.75 \pm 1.29$ & \tabsecond{48.94 $\pm$ 0.44} & \tabsecond{24.32 $\pm$ 0.24} & \tabsecond{29.06 $\pm$ 0.32} & $38.58 \pm 0.39$ \\
& \textbf{Ours} & This work & \tabsecond{73.68 $\pm$ 0.88} & \tabsecond{17.73 $\pm$ 0.88} & \tabbest{53.27 $\pm$ 0.41} & \tabbest{10.18 $\pm$ 0.38} & \tabbest{32.24 $\pm$ 0.05} & \tabbest{22.47 $\pm$ 0.13} \\
\midrule
\multirow{12}{*}{Task-IL} & ER~\citep{riemer2019learning} & ICLR 2019 & $93.61 \pm 0.27$ & $3.50 \pm 0.53$ & $73.98 \pm 1.52$ & $16.23 \pm 1.06$ & $48.64 \pm 0.46$ & $30.73 \pm 0.62$ \\
& iCaRL~\citep{rebuffi2017icarl} & CVPR 2017 & $88.22 \pm 2.62$ & $22.61 \pm 3.97$ & $58.16 \pm 1.76$ & $27.90 \pm 1.37$ & $31.55 \pm 3.27$ & $39.44 \pm 0.84$ \\
& DER~\citep{buzzega2020dark} & NeurIPS 2020 & $93.40 \pm 0.21$ & $3.72 \pm 0.55$ & $71.73 \pm 0.74$ & $25.98 \pm 1.55$ & $51.78 \pm 0.88$ & $28.21 \pm 0.97$ \\
& DER++~\citep{buzzega2020dark} & NeurIPS 2020 & $93.88 \pm 0.50$ & $4.66 \pm 1.15$ & $74.77 \pm 0.31$ & $15.75 \pm 0.48$ & $51.91 \pm 0.68$ & $25.47 \pm 1.03$ \\
& Co$^2$L~\citep{cha2021co2l} & ICCV 2021 & $89.51 \pm 2.45$ & $2.65 \pm 1.00$ & $62.32 \pm 0.19$ & $23.47 \pm 0.27$ & $50.74 \pm 1.24$ & $13.80 \pm 2.10$ \\
& CILA~\citep{wen2024provable} & ICML 2024 & $93.29 \pm 0.24$ & $0.65 \pm 0.13$ & $68.29 \pm 0.46$ & $16.98 \pm 0.56$ & $49.19 \pm 0.91$ & $27.83 \pm 1.32$ \\
& MNC\textsuperscript{3}L~\citep{dang2025memory} & WACV 2025 & $85.94 \pm 0.22$ & $3.15 \pm 0.42$ & $56.43 \pm 0.29$ & $24.29 \pm 0.50$ & $36.32 \pm 0.05$ & $39.00 \pm 0.17$ \\
& STAR~\citep{eskandarstar} & ICLR 2025 & $95.36 \pm 0.37$ & $2.38 \pm 0.34$ & $74.06 \pm 1.63$ & $15.22 \pm 0.96$ & $55.06 \pm 2.07$ & $15.19 \pm 2.61$ \\
& CSReL~\citep{tong2025coreset} & ICLR 2025 & $73.30 \pm 5.92$ & $16.89 \pm 3.19$ & $75.09 \pm 0.78$ & $15.55 \pm 0.76$ & $51.94 \pm 0.22$ & $28.20 \pm 0.36$ \\
& NCT~\citep{NCT1} & ICLR 2023 & $81.27 \pm 0.24$ & $3.84 \pm 0.35$ & $76.06 \pm 0.52$ & \tabsecond{3.97 $\pm$ 0.75} & $62.30 \pm 0.41$ & \tabbest{6.84 $\pm$ 0.87} \\
& ProNC~\citep{wang2026proNC} & ICLR 2026 & \tabbest{96.95 $\pm$ 0.14} & \tabbest{0.24 $\pm$ 0.14} & \tabsecond{86.38 $\pm$ 0.43} & $4.35 \pm 0.17$ & \tabsecond{69.77 $\pm$ 0.89} & $9.52 \pm 0.96$ \\
& \textbf{Ours} & This work & \tabsecond{96.81 $\pm$ 0.04} & \tabsecond{0.54 $\pm$ 0.08} & \tabbest{87.74 $\pm$ 0.07} & \tabbest{2.02 $\pm$ 0.16} & \tabbest{70.16 $\pm$ 0.20} & \tabsecond{7.90 $\pm$ 0.16} \\
\bottomrule
\end{tabular}%
}
\end{table*}

%% file: tables_tex/component_ablation_compact.tex
\begin{table}[H]
  \centering
  \caption{\textbf{Each component targets routing.}
  Seq-CIFAR-100/200 ablation on the same checkpoint features.}
  \label{tab:component_ablation_compact}
  \tiny
  \setlength{\tabcolsep}{4.0pt}
  \begin{tabular}{lccccc rr}
    \toprule
    Head &
    \diaghead{ETF} &
    \diaghead{Transport} &
    \diaghead{Radial} &
    \diaghead{Comp. $\rho_c$} &
    \diaghead{NC shrink} &
    Class-IL $\uparrow$ & Route $\uparrow$ \\
    \midrule
    ETF only
    & \cmark & & & &
    & $41.71\pm0.61$ & $42.81\pm0.57$ \\
    Radial only
    & & \cmark & \cmark & \cmark &
    & $46.21\pm0.44$ & $48.36\pm0.49$ \\
    No transport
    & \cmark & & \cmark & \cmark &
    & $41.71\pm0.61$ & $42.81\pm0.57$ \\
    Equal radius
    & \cmark & \cmark & \cmark & &
    & $44.49\pm0.29$ & $46.04\pm0.28$ \\
    No NC shrink
    & \cmark & \cmark & \cmark & \cmark &
    & $49.20\pm0.48$ & $50.93\pm0.37$ \\
    \textbf{Ours (\method)}
    & \cmark & \cmark & \cmark & \cmark & \cmark
    & $\mathbf{49.74\pm0.58}$ & $\mathbf{51.43\pm0.47}$ \\
    \bottomrule
  \end{tabular}
  \tabnote{Columns correspond to ETF evidence (\cref{eq:etf_score}), transport (\cref{eq:prototype_transport,eq:transported_prototype}), radial evidence (\cref{eq:radial_score}), competitive radii (\cref{eq:competitive_radius}), and NC shrinkage (\cref{eq:nc_shrink_proto}).}
\end{table}

%% file: tables_tex/replay_calibration_main.tex
\begin{table}[H]
  \centering
  \caption{\textbf{Replay-fitted calibration does not explain RCR.}
  A task-bias calibration fit on replay labels is evaluated on held-out
  Seq-CIFAR-100/200 test features at the final task. It worsens routing,
  confirming that the replay buffer is not a reliable calibration set.}
  \label{tab:replay_calibration_main}
  \scriptsize
  \setlength{\tabcolsep}{4.0pt}
  \begin{tabular}{lrrrr}
    \toprule
    Head & Class-IL $\uparrow$ & Task-IL $\uparrow$ & Route $\uparrow$ & Wrong-task $\downarrow$ \\
    \midrule
    ProNC ETF & $41.71\pm0.61$ & $85.53\pm0.16$ & $42.81\pm0.57$ & $57.19\pm0.57$ \\
    Replay task-bias & $29.91\pm0.64$ & $85.53\pm0.16$ & $30.66\pm0.59$ & $69.34\pm0.59$ \\
    \textbf{Ours (\method)} & $\mathbf{49.74\pm0.58}$ & $\mathbf{87.28\pm0.20}$ & $\mathbf{51.43\pm0.47}$ & $\mathbf{48.57\pm0.47}$ \\
    \bottomrule
  \end{tabular}
\end{table}

%% file: tables_tex/nc_cross_dataset_diag.tex
\begin{table}[H]
  \centering
  \caption{\textbf{\method is most active when routing drift is largest.}
  In the exact NC limit, transported class representatives coincide with ETF vertices and \method reduces to nearest-ETF inference. Across validated runs, the head is least active when the ProNC task-local classifier already routes well, and its gains grow with the held-out Task-IL--Route gap (row-wise $r=0.97$ with $\Delta$Route and $r=0.96$ with $\Delta$Class). Values are final-task mean $\pm$ std over three seeds.}
  \label{tab:nc_cross_dataset_diag}
  \scriptsize
  \setlength{\tabcolsep}{2.5pt}
  \begin{tabular}{llrrrrrr}
    \toprule
    Dataset & Buf. & $\bar\delta$ & NC1 disp. & CV$(\rho)$ & Task--Route gap & $\Delta$Route & $\Delta$Class \\
    \midrule
    Seq-CIFAR-10 & 200 & $1.50\pm0.01$ & $0.37\pm0.02$ & $0.13\pm0.00$ & $30.16\pm1.24$ & $+4.23\pm1.41$ & $+4.22\pm1.40$ \\
    Seq-CIFAR-10 & 500 & $1.44\pm0.01$ & $0.38\pm0.02$ & $0.14\pm0.01$ & $24.11\pm1.36$ & $+1.62\pm1.14$ & $+1.52\pm1.05$ \\
    Seq-CIFAR-100 & 200 & $1.61\pm0.00$ & $0.41\pm0.00$ & $0.11\pm0.00$ & $42.71\pm0.41$ & $+8.62\pm0.11$ & $+8.03\pm0.19$ \\
    Seq-CIFAR-100 & 500 & $1.52\pm0.00$ & $0.43\pm0.00$ & $0.10\pm0.00$ & $35.41\pm0.47$ & $+4.90\pm0.32$ & $+4.62\pm0.40$ \\
    Seq-TinyImageNet & 200 & $1.63\pm0.00$ & $0.43\pm0.00$ & $0.08\pm0.00$ & $39.72\pm0.65$ & $+6.18\pm0.72$ & $+5.65\pm0.69$ \\
    Seq-TinyImageNet & 500 & $1.60\pm0.00$ & $0.43\pm0.00$ & $0.07\pm0.00$ & $38.33\pm0.73$ & $+6.09\pm0.36$ & $+5.63\pm0.35$ \\
    \bottomrule
  \end{tabular}
  \tabnote{$\bar\delta$ is the mean cosine distance $d_{\cos}(p_c,u_c)=2-2\langle p_c,u_c\rangle$ between transported class representatives and ETF vertices. Task--Route gap is ProNC Task-IL FAA minus ProNC route accuracy.}
\end{table}

%% file: sections/discussion.tex
\section{Discussion}

\method targets the route-limited regime: the representation remains useful
inside the correct task block, but the global score no longer compares task
blocks reliably. It is not a universal repair for arbitrary forgetting; if
Task-IL is low, the within-block class structure has already been lost. Its
classifier-level scope is useful because the backbone and ETF frame stay fixed:
the gains show that trained progressive NC models contain class-measure
information that the raw ETF score underuses. \method changes class evidence,
not the evaluation protocol, and does not insert a learned task oracle.

This also clarifies the role of \method in the paper. The contribution is not a
large auxiliary learner on top of ProNC, but a controlled test of whether the
NC geometry already learned by the model is being scored correctly. If the raw
ETF classifier were already using all useful geometric information, transported
class centers and radii would not improve held-out routing. The observed gains
therefore indicate a specific mismatch: the global ETF frame remains valuable,
but the local class measure has drifted away from the ideal point prototype.

Replay is used only to estimate class-local concentration around transported
representatives. This differs from fitting a replay task bias, which has label
access but transfers poorly to held-out routing. Inference adds one
prototype-distance score matrix, per-sample standardization, and no extra
backbone forward; for TinyImageNet, the class-level state is about $0.39$ MB in
FP32. The route-oracle analysis in the supplement should therefore be read
constructively: it does not assume a deployable task oracle, but measures how
much of the remaining error is still attributable to block selection rather than
within-block discrimination. The remaining gap motivates training-time
objectives that make representations route-stable rather than only corrected at
inference.

%% file: sections/conclusion.tex
\section{Conclusion}

We isolate routing drift as a measurable failure mode of progressive
neural-collapse continual learning: Task-IL can remain high while Class-IL
fails because the global argmax routes samples to the wrong task block. \method
addresses this regime by preserving the trained ETF frame and adding
transported radial reliability, improving Class-IL and forgetting while
preserving Task-IL. Across the validated settings, the same fixed correction
sets the best Class-IL FAA in five of six cases and lowers forgetting in all
six. This positions neural collapse as both a training target and a diagnostic
geometry for score-comparability failures.

%% file: sections/appendix.tex
\appendix

\section{Supplement Overview}
\label{app:overview}

This supplement is organized as a companion to the main paper. We first record
the implementation details needed to reproduce the trained artifacts, including
the reference ProNC objective and hyperparameters. We then collect notation and
neural-collapse background, and define the classifier variants evaluated from
the same frozen features and saved task artifacts. The remaining sections
separate three questions: which variant is strongest, which components matter,
and whether the observed gains match the routing-drift mechanism. Throughout,
\method denotes the deployed Reliability-Calibrated Routing score; other heads
are controlled variants or ablations.

\section{Implementation Details}
\label{app:implementation_details}

This section records the training and evaluation details behind the frozen
artifacts used in the main paper. Our contribution is evaluated as a classifier
correction on top of a trained progressive ETF model; the backbone checkpoints
are produced with the reference ProNC objective~\cite{wang2026proNC}, while
\method is applied after training from saved task artifacts.

\paragraph{Reference training objective.}
The first task is trained with ordinary supervised cross-entropy. At later
tasks, the ETF frame is expanded before learning the new task, and each update
uses a minibatch containing current-task samples and replay samples. Let
$z_t(x)=f_{\theta_t}(x)/\|f_{\theta_t}(x)\|$ be the normalized current feature,
$z_{t-1}(x)$ the normalized feature from the frozen previous model, and
$u_{y,t}$ the progressive ETF vertex assigned to label $y$ at stage $t$. The
training loss is
\begin{equation}
\mathcal L_{\mathrm{ProNC}}
=
\mathcal L_{\mathrm{ce}}
+\lambda_1\mathcal L_{\mathrm{align}}
+\lambda_2\mathcal L_{\mathrm{distill}}.
\label{eq:app_pronc_objective}
\end{equation}
The alignment term is the squared cosine distance-to-one objective
\begin{equation}
\mathcal L_{\mathrm{align}}(x,y)
=
\frac{1}{2}\left(\inner{z_t(x)}{u_{y,t}}-1\right)^2,
\label{eq:app_align_loss}
\end{equation}
which pulls the current normalized feature toward the ETF vertex of its class.
The feature-distillation term is
\begin{equation}
\mathcal L_{\mathrm{distill}}(x)
=
\frac{1}{2}\left(\inner{z_t(x)}{z_{t-1}(x)}-1\right)^2,
\label{eq:app_distill_loss}
\end{equation}
evaluated on the same mixed minibatch. Thus, $\lambda_1$ controls the strength
of ETF alignment and $\lambda_2$ controls the strength of feature preservation
from the previous model. The cross-entropy term keeps its unit coefficient in
all reproduced runs.

\paragraph{Training hyperparameters.}
All reproduced artifacts use a ResNet-18 backbone, batch size $32$, SGD with
zero momentum and zero weight decay, and a multistep learning-rate schedule
with decay factor $0.1$. The class order is fixed by the benchmark split, and
all reported paired results average seeds $0,1,2$. The dataset-specific values
used to produce the saved checkpoints are listed in
\cref{tab:implementation_hparams}.

\begin{table}[t]
\centering
\caption{\textbf{Training hyperparameters for reproduced artifacts.}
The same checkpoints are used to evaluate both the reference ETF head and
\method in paired comparisons.}
\label{tab:implementation_hparams}
\scriptsize
\setlength{\tabcolsep}{4pt}
\begin{tabular}{@{}lcccccc@{}}
\toprule
Dataset & Buffer & Epochs & LR & Milestones & $\lambda_1$ & $\lambda_2$ \\
\midrule
Seq-CIFAR-10 & 200 & 50 & 0.01 & 35, 45 & 13 & 90 \\
Seq-CIFAR-10 & 500 & 50 & 0.01 & 35, 45 & 12 & 80 \\
Seq-CIFAR-100 & 200 & 50 & 0.03 & 35, 45 & 18 & 170 \\
Seq-CIFAR-100 & 500 & 50 & 0.03 & 35, 45 & 18 & 170 \\
Seq-TinyImageNet & 200 & 100 & 0.03 & 45, 75, 95 & 20 & 165 \\
Seq-TinyImageNet & 500 & 100 & 0.03 & 45, 75, 95 & 20 & 165 \\
\bottomrule
\end{tabular}
\end{table}

\paragraph{Saved artifacts and post-hoc evaluation.}
At each task boundary we save the checkpoint, the expanded ETF frame, replay
features needed for class representatives, and the previous-to-current feature
transport fit. The reference ETF head predicts with
\begin{equation}
\hat y_{\mathrm{ETF}}(x)=
\arg\max_{c\in\mathcal C_t}\inner{z_t(x)}{u_{c,t}}.
\end{equation}
\method uses the same frozen feature $z_t(x)$ and the same ETF frame. It adds
only the transported class representative, competitive radius, NC retraction,
and fixed ETF/radial composition defined in the main paper. No scalar is chosen
per dataset or from a validation split: all datasets use $\alpha_\rho=0.25$,
$\gamma=0.5$, shrink clipping to $[0,0.95]$, radius clipping to
$[10^{-4},10]$, and $\omega=q/(q+1)$.

\paragraph{Transport fitting.}
Class representatives are moved across feature frames by fitting a ridge
linear map between previous-model and current-model normalized features from
the mixed training stream at a task boundary. The reported artifacts use ridge
coefficient $10^{-3}$, pair normalization before fitting, and normalized
features at evaluation. The transported representative $p_c$ is therefore a
class-level measure in the current feature frame, while $u_c$ remains the
global ETF vertex used by the neural-collapse classifier.

\section{Notation and Neural-Collapse Background}
\label{app:neural_collapse}

\Cref{tab:notation} summarizes the notation used throughout the paper and this
supplement. We keep protocol symbols, routing diagnostics, and \method
components separate to avoid overloading task indices, class sets, and prototype
quantities.

\input{tables_tex/notation}
\FloatBarrier

We recall the standard neural-collapse terminology used in the main paper.
Consider a balanced $K$-class classification problem. Let
$h_{k,i}\in\mathbb{R}^d$ be the last-layer feature of sample $i$ from class $k$,
$\mu_k$ the class mean, $\mu_G=\frac{1}{K}\sum_{k=1}^{K}\mu_k$ the global mean,
and $w_k$ the classifier weight for class $k$. Neural collapse describes the
terminal training regime in which the following four properties
emerge~\cite{papyan2020prevalence}.

\paragraph{NC1: within-class variability collapse.}
Features from the same class concentrate around their class mean:
\begin{equation}
\Sigma_W=\frac{1}{K}\sum_{k=1}^{K}\frac{1}{n_k}\sum_i
(h_{k,i}-\mu_k)(h_{k,i}-\mu_k)^\top \rightarrow 0.
\end{equation}

\paragraph{NC2: simplex ETF structure of class means.}
The centered class means
\begin{equation}
\tilde{\mu}_k=\frac{\mu_k-\mu_G}{\|\mu_k-\mu_G\|}
\end{equation}
approach a simplex equiangular tight frame, so all classes have equal norm and
equal pairwise angle:
\begin{equation}
\inner{\tilde{\mu}_i}{\tilde{\mu}_j}=-\frac{1}{K-1}
\qquad i\neq j.
\end{equation}

\paragraph{NC3: self-duality.}
Classifier weights align with centered class means:
\begin{equation}
\frac{w_k}{\|w_k\|}\approx \tilde{\mu}_k.
\end{equation}

\paragraph{NC4: nearest-class-center inference.}
Under NC1--NC3, the classifier decision is equivalent to nearest-class-center
or nearest-prototype inference in the collapsed feature space. This is the
connection exploited by ETF classifiers: once the class directions form a
simplex ETF, prediction can be performed by comparing a normalized feature to
class vertices through cosine similarity.

\section{Offline Classifier Variants}
\label{app:offline_variants}
\label{app:offline_heads}

All variants below are evaluated on the same frozen feature tensor for a given
run. We use $z$ for the normalized feature, $u_c$ for the ProNC ETF vertex of
class $c$, and $p_c$ for the transported class prototype in the current feature
frame:
\begin{equation}
z=\frac{f_\theta(x)}{\|f_\theta(x)\|}.
\end{equation}
We write $\mathcal N[\cdot]$ for per-sample standardization across the currently
seen classes. Unless stated otherwise, no variant uses test-time task identity
or a learned task oracle.

\paragraph{Evaluation artifacts.}
All variants are evaluated from saved task and feature artifacts. The same saved
features are used for ProNC ETF, LDC-NCM, radial scoring, and
reliability-calibrated scoring. This isolates the classifier effect from
training variance.

\paragraph{ETF reference.}
This is the nearest-ETF score used by the base progressive neural-collapse
classifier:
\begin{equation}
\scoreetf(x,c)=\inner{z}{u_c}.
\end{equation}
This is the baseline head in the main tables.

\paragraph{CE head (audit).}
The trained linear classifier head uses its saved weights and bias. We report
it only as an audit row, because the controlled setting evaluates
neural-collapse heads with ETF inference rather than this linear classifier.

\paragraph{Transported NCM.}
This variant uses transported nearest-centroid inference:
\begin{equation}
\scoreldc(x,c)=\inner{z}{p_c}.
\end{equation}
It tests whether moving class representatives into the current feature frame is
useful without adding radial reliability.

\paragraph{Radial-only.}
This variant scores local plausibility by normalizing the cosine distance to
$p_c$ with a class radius:
\begin{equation}
\scorerad(x,c)=
-\frac{2-2\inner{z}{p_c}}{\rho_c+\epsilon}.
\end{equation}
It tests whether local class reliability alone can repair cross-task routing.

\paragraph{CompMix.}
CompMix composes the ETF score with the competitive-radius radial score around
the transported prototype:
\begin{equation}
\scorecomp(x,c)
=(1-\lambda_t)\mathcal N[\scoreetf](x,c)
+\lambda_t\mathcal N[\scorerad](x,c),
\qquad
\lambda_t=\min\left(\frac{q}{q+1},\frac{t-1}{t}\right).
\end{equation}
The deterministic stage schedule keeps the radial term weak at early tasks and
approaches the fixed cap as more task blocks are observed.

\paragraph{CompMix-Fixed.}
CompMix-Fixed uses the same competitive-radius radial score around $p_c$, but
removes the stage schedule and uses the fixed cap $\omega=q/(q+1)$ at every
task:
\begin{equation}
\scorecompfix(x,c)
=(1-\omega)\mathcal N[\scoreetf](x,c)
+\omega\mathcal N[\scorerad](x,c).
\end{equation}
It isolates whether the gain comes from the evidence composition itself or from
a deterministic task-stage schedule.

\paragraph{InvGradMix (directional sensitivity).}
InvGradMix replaces the Euclidean radial distance with a global diagonal metric
$M_t=\operatorname{diag}(m)$ estimated from replay samples using inverse
empirical CE-gradient magnitude:
\begin{equation}
\scoreinvgrad(x,c)=
-\frac{(z-p_c)^\top M_t(z-p_c)}{\rho_c+\epsilon}.
\end{equation}
It then uses the same stage schedule $\lambda_t$ as CompMix. This variant tests
whether different feature directions should have different reliability.

\paragraph{InvGradMix-Fixed (directional sensitivity).}
InvGradMix-Fixed uses the same direction-aware radial score as InvGradMix, but
uses the fixed cap $\omega=q/(q+1)$ instead of the stage schedule. It is a
stronger direction-aware variant, but it introduces metric-estimation choices
that are not needed by \method.

\paragraph{NC-ShrinkMix.}
NC-ShrinkMix adds the neural-collapse retraction to CompMix. It first retracts
the transported prototype toward the ETF vertex,
\begin{equation}
\tilde p_c=\operatorname{norm}\left(\alpha_c p_c+(1-\alpha_c)u_c\right),
\qquad
\alpha_c=
\operatorname{clip}\left[
\left(\frac{\bar\rho}{\rho_c+\epsilon}\right)^\gamma,
0,0.95
\right],
\label{eq:app_nc_shrink_proto}
\end{equation}
with $\gamma=0.5$. It then recomputes the competitive class radius $\rho_c$ and
combines ETF evidence with radial evidence using the scheduled mixture
$\lambda_t$.

\paragraph{RCR.}
\method is the fixed-cap version of NC-ShrinkMix. It uses the retracted
prototype $\tilde p_c$, recomputes the competitive radius around that prototype,
and combines ETF evidence with radial evidence using $\omega=q/(q+1)$:
\begin{equation}
\scorercr(x,c)
=(1-\omega)\mathcal N[\scoreetf](x,c)
+\omega\mathcal N[\scorerad](x,c).
\label{eq:app_rcr_score}
\end{equation}
This is the method used in all main-paper results.

\paragraph{Saved mix (audit).}
Some artifacts contain the mixture that was active when the run was produced.
Because this score can depend on historical choices of auxiliary head, radius
mode, and mixture weight, we report it only as an audit row. It is not a method
candidate and is included only to make older saved artifacts auditable.

\paragraph{No class-index prior.}
The proposed score does not use the numerical class index, task age, or a
hand-coded penalty such as $a_{\tau(c)}$ or $b\,c$ in the logits. The only
mixture quantity in \method is the global scalar $\omega=q/(q+1)$, shared by
every class and sample. Class-specific terms are data-derived representatives
and uncertainties, $(u_c,p_c,\rho_c)$, exactly as in prototype-based
classifiers. The task index is used for evaluation metrics such as Task-IL and
route accuracy, but not as a test-time oracle or as a per-class routing bias.

\section{Classifier Variant Benchmark}
\label{app:variant_benchmark}

The main paper reports the paired ETF-versus-\method comparison. Here we show
the broader classifier family on the same frozen artifacts. The purpose is not
to introduce multiple methods, but to verify which ingredient explains the
gains. RCR is the deployed variant because it combines ETF evidence, transported
class reliability, and NC retraction without the additional metric-estimation
choices used by the direction-aware variants. Rows explicitly marked as audits
are diagnostic controls rather than candidate methods; they are included to show
which apparent shortcuts fail under the same frozen-feature evaluation.

\paragraph{Analysis.}
\Cref{tab:variant_benchmark} shows three consistent patterns. First, the
audit CE head is not a competitive replacement for ETF inference: it often
preserves Task-IL but collapses Class-IL, confirming that the comparison should
be made between neural-collapse heads on the same feature artifacts. Second,
transported NCM and radial-only scoring are useful but incomplete: they improve
hard routing regimes such as Seq-CIFAR-100 and Seq-TinyImageNet, yet either
underperform on benign Seq-CIFAR-10 or lose the global ETF ordering. Third, the
large gains appear when the ETF score and local radial reliability are composed.
CompMix and NC-ShrinkMix therefore separate the two useful signals: global ETF
comparability and local class plausibility. The fixed variants remove the
deterministic stage schedule; RCR is selected because it gives the most stable
forgetting reduction while keeping the isotropic NC story intact. The
direction-aware InvGradMix variants are deliberately left as ablations: they
show that additional feature-direction reliability can sometimes improve FAA or
Route, but at the cost of an extra learned metric and a less parsimonious
method.

\input{tables_tex/variant_benchmark}
\FloatBarrier

\section{Second Progressive-ETF Diagnostic}
\label{app:nct_diagnostic}

Reviewer feedback asked whether the route-correction mechanism is tied to one
progressive-ETF implementation. To address this without overclaiming, we ran a
small diagnostic on an NCT-style progressive ETF model and applied an offline
RCR-style ramp correction to the same frozen features. This is not intended as
a full NCT reproduction or a replacement for the controlled ProNC comparison in
the main paper. It simply checks whether the same route-repair signal appears
when the base learner is changed within the NC/ETF family.

\input{tables_tex/nct_diagnostic}
\FloatBarrier

\section{Component and Sensitivity Ablations}
\label{app:component_sensitivity}

\subsection{Score Components}
\label{app:score_components}

\input{tables_tex/score_component_ablation}
\FloatBarrier

\paragraph{Analysis.}
\Cref{tab:score_component_ablation} uses the same frozen evaluation artifacts
as the main results and separates ETF evidence, transported centroids, radial
evidence, competitive radii, and transport. It reveals two failure modes.
Transported NCM usually improves over ETF on hard routing regimes, so class
representatives do benefit from being moved into the current feature frame.
However, NCM alone does not supply a reliability model and can underperform on
easier regimes. Conversely, ETF+no-transport collapses back to the ETF
reference, showing that radial evidence only matters when it is attached to
transported class measures. The full RCR row is strongest because it keeps the
global ETF ordering while adding local class plausibility and competitive
uncertainty.

\subsection{Mixture Weight and Schedule}
\label{app:mixture_weight_schedule}

\Cref{tab:omega_sensitivity} shows that the fixed $q/(q+1)$ mixture is in a
broad useful region rather than a hidden test-selected constant.
\Cref{tab:rcr_schedule_ablation} tests whether the deterministic stage schedule
is necessary for the Seq-CIFAR-100 artifacts. The fixed-cap variant removes the
stage ramp and is used by the main paper.

\input{tables_tex/omega_sensitivity}
\input{tables_tex/rcr_schedule_ablation}
\FloatBarrier

\paragraph{Analysis.}
The mixture ablations rule out a narrow schedule artifact. A broad interval of
$\omega$ remains competitive, and removing the deterministic stage schedule
does not remove the gain. This is why the main method uses the simpler fixed
cap $\omega=q/(q+1)$: it is tied to the number of classes per block rather than
to a tuned task index, while preserving the Class-IL and route improvements.
Even the global value $\omega=0.5$ improves over ETF in every dataset/buffer
cell in \cref{tab:omega_sensitivity}, so the gain is not created by a single
test-selected mixture constant.

\subsection{Radius and NC Retraction}
\label{app:radius_shrinkage}

\Cref{tab:alpha_rho_sensitivity} shows that the competitive radius exponent is
not a narrow tuned constant; the default $\alpha_\rho=0.25$ lies in a stable
range. \Cref{tab:shrink_clip_ablation} isolates the NC-retraction clipping: the
lower endpoint is the natural value $0$, while the upper safeguard $0.95$ avoids
fully trusting transported centers after drift compensation.

\input{tables_tex/alpha_rho_sensitivity}
\input{tables_tex/shrink_clip_ablation}
\FloatBarrier

\paragraph{Analysis.}
The radius and shrinkage tables show that RCR is not driven by a fragile radius
constant. Moderate changes in $\alpha_\rho$ keep the same qualitative behavior:
larger class uncertainty reduces radial confidence, and the ETF component
preserves global comparability. The NC retraction mainly stabilizes forgetting
and late-stage routing by preventing unreliable transported centers from moving
too far away from their neural-collapse vertices.

\section{Replay Calibration Baselines}
\label{app:replay_calibration}

Replay-fitted calibrators are useful stress tests because they fit task or class
biases from replay labels. They are not the same claim as \method: RCR does not
fit a task router from replay and does not add a task oracle. The comparison in
\cref{tab:calibration_baselines_all} asks whether replay-side score calibration
alone explains the gains. It does not: replay calibration can help in some
TinyImageNet cells, but it is not consistently competitive with RCR and can
damage CIFAR routing.

\input{tables_tex/calibration_baselines_all}
\FloatBarrier

\paragraph{Analysis.}
Replay calibration is a useful negative control. If routing drift were only a
global task-bias problem, replay-fitted offsets would match RCR. Instead, their
behavior is inconsistent across datasets and memory budgets. This supports the
main interpretation: the useful signal is not a learned task router, but the
combination of transported class evidence and class-specific reliability under
the same ETF decision space.

\section{Mechanistic Diagnostics}
\label{app:mechanistic_diagnostics}

The main paper argues that RCR acts when learned class measures deviate from the
symmetric ETF geometry. The diagnostics below make this claim explicit through
task-by-task curves, score-profile examples, stage-level NC proxies, route
oracle bounds, and class-wise correlations.

\begin{figure}[t]
  \centering
  \includegraphics[width=\linewidth]{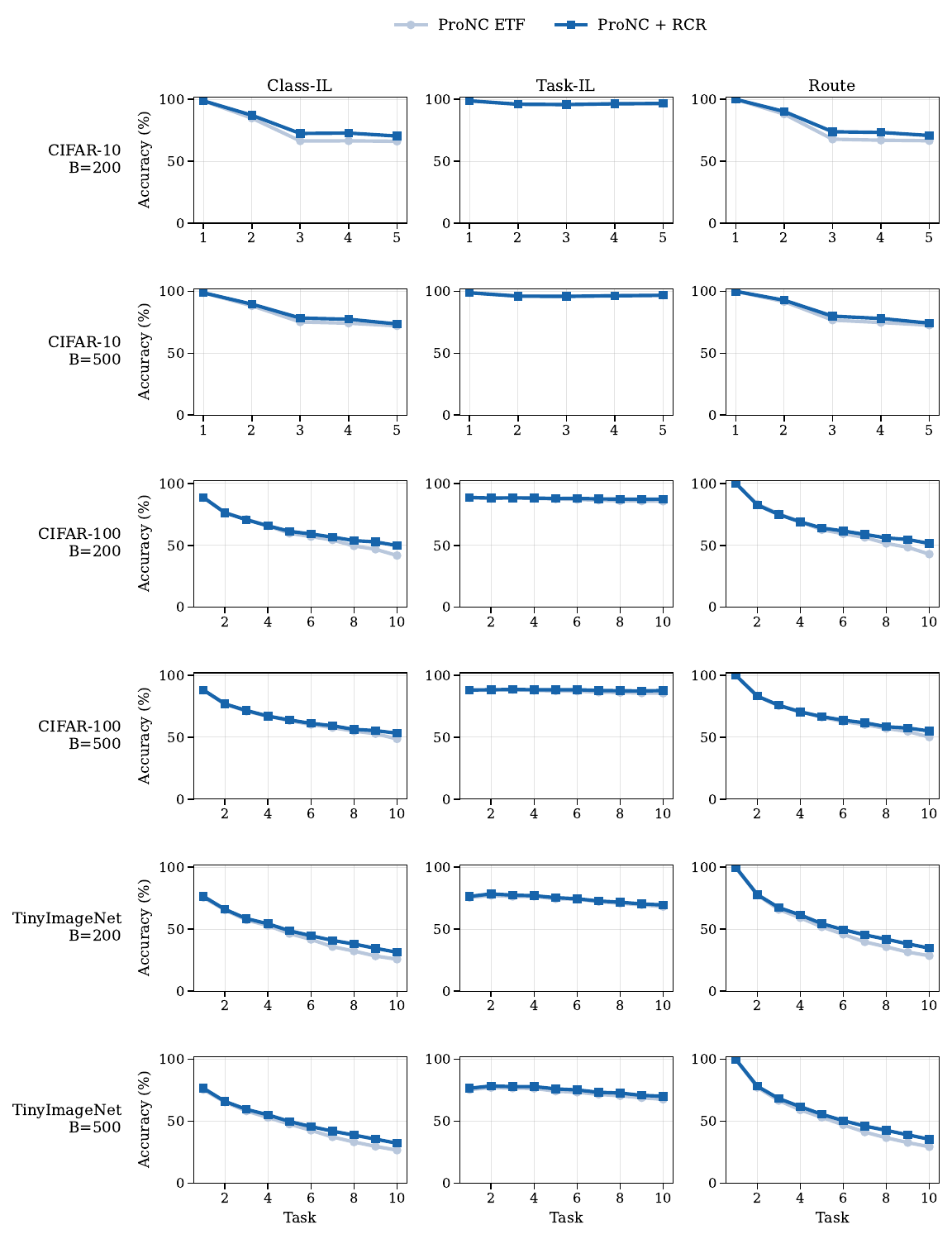}
  \caption{\textbf{Task-by-task diagnostic curves across all validated datasets.}
  Each row shows one dataset and memory budget; columns report Class-IL accuracy, Task-IL accuracy, and route accuracy over the continual sequence. The same frozen checkpoints are evaluated with the ProNC ETF head and \methodhead. Across datasets, \method primarily changes the Class-IL and Route curves while leaving Task-IL close to the reference head, supporting the routing-drift interpretation.}
  \label{fig:appendix_all_diagnostic_curves}
\end{figure}

\begin{figure}[t]
  \centering
  \includegraphics[width=\linewidth]{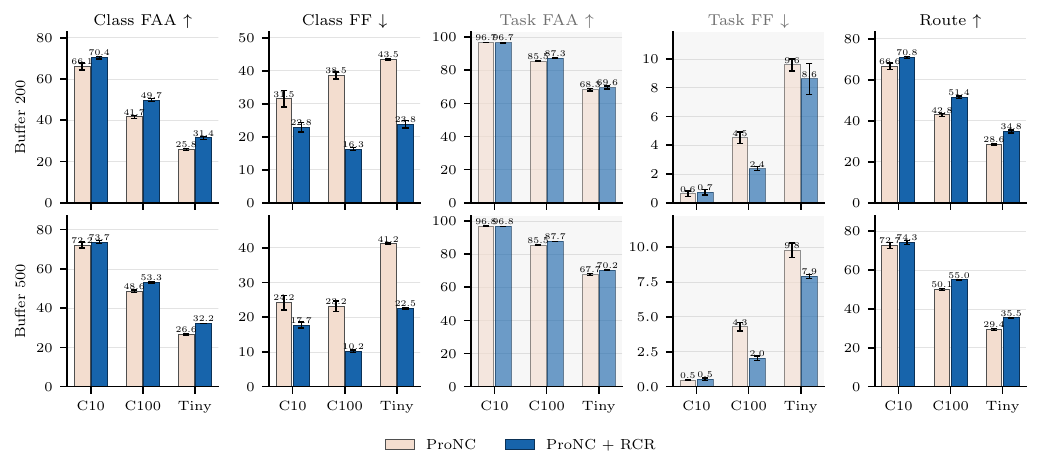}
  \caption{\textbf{Final head comparison across datasets and memory budgets.}
  Each panel compares the ProNC ETF head with \probase+\method on the same
  saved evaluation artifacts. Rows correspond to buffer sizes 200 and 500; columns
  report Class-IL FAA, Class-IL FF, Task-IL FAA, Task-IL FF, and final route
  accuracy.}
  \label{fig:appendix_final_metric_grid}
\end{figure}

\begin{figure}[t]
  \centering
  \includegraphics[width=0.78\linewidth]{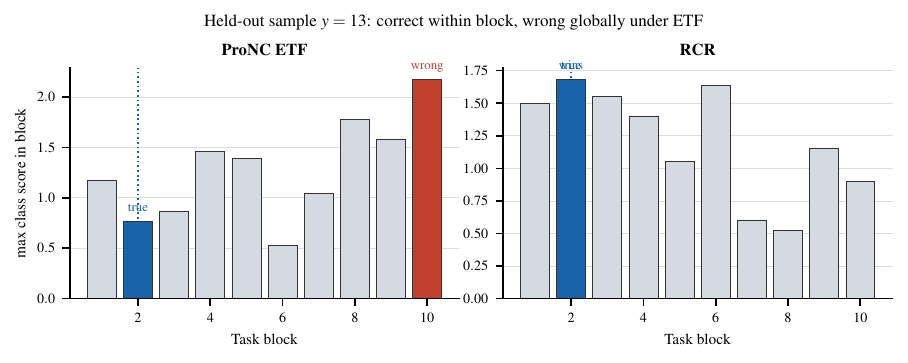}
  \caption{\textbf{A routing error at the block-score level.}
  For one held-out Seq-CIFAR-100 sample, the true class wins within its own
  task block, but the ProNC ETF maximum is larger in a wrong block. \method
  changes the cross-block score profile so the correct block wins, without
  using task identity at test time.}
  \label{fig:appendix_task_block_score_profile}
\end{figure}
\FloatBarrier

\paragraph{Analysis.}
The diagnostic curves and block-score example visualize the same mechanism as
the tables. RCR changes Class-IL and route accuracy together, while Task-IL is
largely preserved. The block-score example makes this concrete: the sample is
not a within-task class failure, because the correct class wins inside its own
block; the error is caused by cross-block score comparability, which the
reliability-calibrated score changes.

\input{tables_tex/nc_deviation_diag}
\input{tables_tex/route_oracle_bounds}
\input{tables_tex/classwise_nc_correlation}

\begin{figure}[t]
  \centering
  \includegraphics[width=0.58\linewidth]{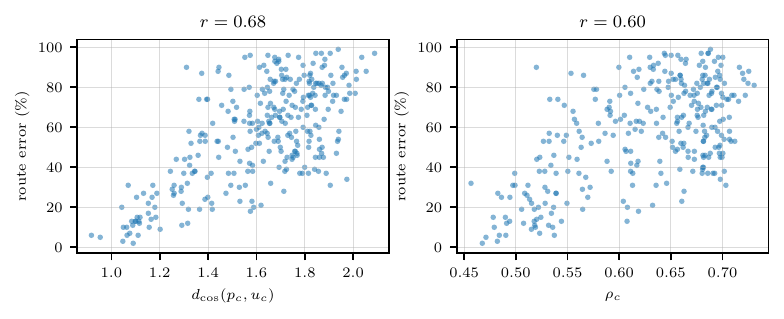}
  \caption{\textbf{Class-wise NC-deviation signals predict routing error.}
  On Seq-CIFAR-100 with buffer 200, classes whose transported prototypes deviate
  further from their ETF vertices, or whose competitive radii are larger, tend
  to have larger held-out route error under the ProNC ETF head.}
  \label{fig:classwise_nc_correlation}
\end{figure}
\FloatBarrier

\paragraph{Analysis.}
\Cref{tab:nc_deviation_diag} reports stage-level NC diagnostics and class-wise
correlations, while \cref{tab:route_oracle_bounds} reports the true-route upper
bound. Together they show that routing, not task-local discrimination, is the
dominant remaining headroom. Finally,
\cref{tab:classwise_nc_correlation,fig:classwise_nc_correlation} connect the
effect to neural-collapse deviations at class level: classes with larger
prototype-to-ETF deviation or larger competitive radius tend to have larger
route error, which is exactly where radial reliability should matter.
\FloatBarrier

%% file: tables_tex/notation.tex
\begin{table}[H]
\centering
\caption{\textbf{Primary notation.}
The first column makes each object's type explicit. We use calligraphic
letters for sets, lower-case symbols for feature/prototype vectors and
scalars, upper-case Roman letters for matrices, and
$\mathcal L_{\mathrm{name}}$ for losses.}
\label{tab:notation}
\scriptsize
\setlength{\tabcolsep}{3pt}
\begin{tabular}{@{}p{0.12\textwidth}p{0.23\textwidth}p{0.57\textwidth}@{}}
\toprule
Type & Symbol & Description \\
\midrule
\multicolumn{3}{@{}l}{\textbf{Protocol and indexing}} \\
Set & $\mathcal T=\{1,\ldots,T\}$ & Ordered continual-learning task sequence. \\
Index & $t$ & Current training or evaluation stage. \\
Set & $\mathcal D_t$ & Training data introduced by task $t$. \\
Set & $\mathcal C_t$ & Classes observed after task $t$. \\
Scalar & $q$ & Number of classes per task block. \\
Function & $\tau(c)$ & Task block containing class $c$. \\
Set & $\mathcal B_c$ & Replay samples stored for class $c$. \\
Metric & FAA, FF & Final average accuracy and final forgetting. \\
\midrule
\multicolumn{3}{@{}l}{\textbf{Features, vectors, and matrices}} \\
Function & $f_\theta$ & Current feature extractor. \\
Vector & $z(x)$ & Normalized current feature, $f_\theta(x)/\|f_\theta(x)\|$. \\
Vector & $z^{-}(x)$ & Normalized feature from the previous model. \\
Vector & $u_c$ & Progressive ETF vertex for class $c$. \\
Matrix & $A_t$ & Ridge previous-to-current feature transport at stage $t$. \\
Vector & $p_c$ & Transported class representative in the current feature frame. \\
Vector & $\tilde p_c$ & NC-shrunk class representative used by the radial score. \\
Operator & $d_{\cos}(a,b)$ & Cosine-distance convention $2-2\langle a,b\rangle$ for normalized vectors. \\
Operator & $\mathcal N[\cdot]$ & Per-sample standardization over currently seen classes. \\
\midrule
\multicolumn{3}{@{}l}{\textbf{Scores, radii, and scalars}} \\
Score & $\scoreetf(x,c)$ & ETF cosine score for feature $z(x)$ and vertex $u_c$. \\
Score & $\scorerad(x,c)$ & Radius-normalized radial score around the NC-shrunk class representative. \\
Score & $\scorercr(x,c)$ & Final \method score combining standardized ETF and radial evidence. \\
Scalar & $r_c$ & Raw class radius estimated from replay features. \\
Scalar & $\bar r$ & Mean raw radius over currently seen classes. \\
Scalar & $\rho_c$ & Competitive class-dependent reliability radius. \\
Scalar & $\bar\rho$ & Mean competitive radius. \\
Scalar & $\alpha_\rho$ & Radius-compression exponent used to form competitive radii. \\
Scalar & $\alpha_c$ & NC-shrink coefficient used to retract $p_c$ toward $u_c$. \\
Scalar & $\gamma$ & Exponent controlling the shrink coefficient. \\
Scalar & $\omega$ & Fixed ETF/radial mixture weight, $\omega=q/(q+1)$. \\
Scalar & $\epsilon$ & Numerical stability constant. \\
\midrule
\multicolumn{3}{@{}l}{\textbf{Losses and routing diagnostics}} \\
Loss & $\mathcal L_{\mathrm{ProNC}}$ & Reference ProNC training objective. \\
Loss & $\mathcal L_{\mathrm{ce}}$ & Supervised cross-entropy term. \\
Loss & $\mathcal L_{\mathrm{align}}$ & ETF-alignment term in the reference objective. \\
Loss & $\mathcal L_{\mathrm{distill}}$ & Feature-distillation term from the previous model. \\
Score & $m_b(x)$ & Maximum class score inside task block $b$. \\
Scalar & $\Delta_{\mathrm{route}}(x)$ & True-block score margin, $m_{\tau(y)}(x)-\max_{b\neq\tau(y)}m_b(x)$. \\
Event & $R(x)$ & Indicator that the global argmax lies in the correct task block. \\
Event & $Q(x)$ & Indicator that the correct class wins inside the true block. \\
Metric & Route accuracy & Test-set expectation of $R(x)$. \\
\bottomrule
\end{tabular}
\end{table}

%% file: tables_tex/variant_benchmark.tex
\begin{table}[H]
\centering
\caption{\textbf{Classifier variants evaluated on frozen artifacts.}
All variants use the same feature artifacts and test split. Route is final route accuracy.}
\label{tab:variant_benchmark}
\tiny
\setlength{\tabcolsep}{2.0pt}
\renewcommand{\arraystretch}{0.86}
\resizebox{0.98\textwidth}{!}{%
\begin{tabular}{lllrrrrrr}
\toprule
Dataset & Buffer & Head & $n$ & Class FAA $\uparrow$ & Class FF $\downarrow$ & Task FAA $\uparrow$ & Task FF $\downarrow$ & Route $\uparrow$ \\
\midrule
\multirow{11}{*}{CIFAR-10} & \multirow{11}{*}{200} & ETF reference & 3 & 66.14 $\pm$ 1.69 & 31.47 $\pm$ 2.46 & \tabbest{96.74 $\pm$ 0.12} & \tabbest{0.62 $\pm$ 0.20} & 66.58 $\pm$ 1.64 \\
 &  & CE head (audit) & 3 & 54.63 $\pm$ 0.15 & 52.99 $\pm$ 0.03 & 96.62 $\pm$ 0.24 & 0.75 $\pm$ 0.24 & 55.04 $\pm$ 0.09 \\
 &  & Transported NCM & 3 & 63.46 $\pm$ 0.75 & \tabsecond{20.75 $\pm$ 1.21} & 96.11 $\pm$ 0.53 & 1.36 $\pm$ 0.66 & 63.98 $\pm$ 0.66 \\
 &  & Radial-only & 3 & 64.38 $\pm$ 1.23 & \tabbest{15.80 $\pm$ 1.84} & 96.11 $\pm$ 0.53 & 1.36 $\pm$ 0.66 & 65.06 $\pm$ 0.98 \\
 &  & CompMix & 3 & 69.84 $\pm$ 0.76 & 24.39 $\pm$ 1.40 & 96.63 $\pm$ 0.10 & 0.75 $\pm$ 0.22 & 70.30 $\pm$ 0.80 \\
 &  & CompMix-Fixed & 3 & 69.84 $\pm$ 0.76 & 24.15 $\pm$ 1.46 & 96.63 $\pm$ 0.10 & 0.77 $\pm$ 0.22 & 70.30 $\pm$ 0.80 \\
 &  & InvGradMix & 3 & \tabbest{70.65 $\pm$ 0.71} & 22.59 $\pm$ 1.17 & \tabsecond{96.72 $\pm$ 0.09} & \tabsecond{0.67 $\pm$ 0.14} & \tabbest{71.11 $\pm$ 0.73} \\
 &  & InvGradMix-Fixed & 3 & 70.65 $\pm$ 0.71 & 22.16 $\pm$ 1.18 & 96.72 $\pm$ 0.09 & 0.69 $\pm$ 0.13 & 71.11 $\pm$ 0.73 \\
 &  & NC-ShrinkMix & 3 & \tabsecond{70.36 $\pm$ 0.62} & 23.03 $\pm$ 1.40 & 96.67 $\pm$ 0.09 & 0.72 $\pm$ 0.19 & \tabsecond{70.82 $\pm$ 0.56} \\
 &  & RCR & 3 & 70.36 $\pm$ 0.62 & 22.85 $\pm$ 1.45 & 96.67 $\pm$ 0.09 & 0.73 $\pm$ 0.19 & 70.82 $\pm$ 0.56 \\
 &  & Saved mix (audit) & 3 & 68.41 $\pm$ 0.58 & 26.72 $\pm$ 0.93 & 96.59 $\pm$ 0.30 & 0.80 $\pm$ 0.38 & 68.93 $\pm$ 0.51 \\
\addlinespace[2pt]
\multirow{11}{*}{CIFAR-10} & \multirow{11}{*}{500} & ETF reference & 3 & 72.16 $\pm$ 1.59 & 24.18 $\pm$ 2.14 & \tabsecond{96.82 $\pm$ 0.14} & \tabbest{0.47 $\pm$ 0.04} & 72.71 $\pm$ 1.65 \\
 &  & CE head (audit) & 3 & 65.60 $\pm$ 0.48 & 38.90 $\pm$ 0.48 & \tabbest{96.86 $\pm$ 0.15} & 0.51 $\pm$ 0.10 & 66.03 $\pm$ 0.48 \\
 &  & Transported NCM & 3 & 66.28 $\pm$ 0.34 & 18.60 $\pm$ 0.53 & 96.34 $\pm$ 0.04 & 0.97 $\pm$ 0.13 & 66.95 $\pm$ 0.36 \\
 &  & Radial-only & 3 & 67.81 $\pm$ 1.05 & \tabbest{12.52 $\pm$ 1.58} & 96.34 $\pm$ 0.04 & 0.97 $\pm$ 0.13 & 68.51 $\pm$ 0.96 \\
 &  & CompMix & 3 & 73.56 $\pm$ 0.04 & 18.42 $\pm$ 0.85 & 96.76 $\pm$ 0.01 & 0.59 $\pm$ 0.12 & 74.22 $\pm$ 0.19 \\
 &  & CompMix-Fixed & 3 & 73.56 $\pm$ 0.04 & 18.35 $\pm$ 0.82 & 96.76 $\pm$ 0.01 & 0.59 $\pm$ 0.13 & 74.22 $\pm$ 0.19 \\
 &  & InvGradMix & 3 & \tabbest{74.00 $\pm$ 0.19} & 17.25 $\pm$ 0.94 & 96.86 $\pm$ 0.09 & \tabsecond{0.48 $\pm$ 0.01} & \tabbest{74.65 $\pm$ 0.32} \\
 &  & InvGradMix-Fixed & 3 & 74.00 $\pm$ 0.19 & \tabsecond{17.06 $\pm$ 1.01} & 96.86 $\pm$ 0.09 & 0.49 $\pm$ 0.02 & 74.65 $\pm$ 0.32 \\
 &  & NC-ShrinkMix & 3 & \tabsecond{73.68 $\pm$ 0.88} & 17.78 $\pm$ 0.90 & 96.81 $\pm$ 0.04 & 0.54 $\pm$ 0.07 & \tabsecond{74.33 $\pm$ 1.02} \\
 &  & RCR & 3 & 73.68 $\pm$ 0.88 & 17.73 $\pm$ 0.88 & 96.81 $\pm$ 0.04 & 0.54 $\pm$ 0.08 & 74.33 $\pm$ 1.02 \\
 &  & Saved mix (audit) & 3 & 72.90 $\pm$ 0.56 & 19.97 $\pm$ 0.76 & 96.79 $\pm$ 0.01 & 0.52 $\pm$ 0.15 & 73.53 $\pm$ 0.49 \\
\addlinespace[2pt]
\multirow{11}{*}{CIFAR-100} & \multirow{11}{*}{200} & ETF reference & 3 & 41.71 $\pm$ 0.75 & 38.55 $\pm$ 1.07 & 85.53 $\pm$ 0.20 & 4.51 $\pm$ 0.40 & 42.81 $\pm$ 0.70 \\
 &  & CE head (audit) & 3 & 19.77 $\pm$ 0.51 & 77.33 $\pm$ 0.88 & 86.32 $\pm$ 0.10 & 3.76 $\pm$ 0.20 & 20.48 $\pm$ 0.51 \\
 &  & Transported NCM & 3 & 43.59 $\pm$ 0.79 & 31.24 $\pm$ 1.00 & 85.64 $\pm$ 0.09 & 4.03 $\pm$ 0.17 & 45.41 $\pm$ 0.78 \\
 &  & Radial-only & 3 & 46.78 $\pm$ 0.79 & 21.47 $\pm$ 0.80 & 85.64 $\pm$ 0.09 & 4.03 $\pm$ 0.17 & 48.92 $\pm$ 0.65 \\
 &  & CompMix & 3 & 49.28 $\pm$ 0.50 & 21.13 $\pm$ 0.46 & 87.04 $\pm$ 0.21 & 2.76 $\pm$ 0.29 & 50.99 $\pm$ 0.38 \\
 &  & CompMix-Fixed & 3 & 49.20 $\pm$ 0.48 & 18.24 $\pm$ 0.47 & 87.00 $\pm$ 0.18 & 2.62 $\pm$ 0.18 & 50.93 $\pm$ 0.37 \\
 &  & InvGradMix & 3 & 49.55 $\pm$ 0.60 & 20.76 $\pm$ 0.68 & 87.17 $\pm$ 0.22 & 2.57 $\pm$ 0.17 & 51.26 $\pm$ 0.50 \\
 &  & InvGradMix-Fixed & 3 & 49.46 $\pm$ 0.51 & \tabsecond{18.00 $\pm$ 0.50} & 87.13 $\pm$ 0.19 & 2.47 $\pm$ 0.17 & 51.21 $\pm$ 0.42 \\
 &  & NC-ShrinkMix & 3 & \tabsecond{49.72 $\pm$ 0.66} & 19.31 $\pm$ 0.62 & \tabbest{87.37 $\pm$ 0.28} & \tabsecond{2.45 $\pm$ 0.22} & \tabsecond{51.38 $\pm$ 0.50} \\
 &  & RCR & 3 & \tabbest{49.74 $\pm$ 0.71} & \tabbest{16.27 $\pm$ 0.45} & \tabsecond{87.28 $\pm$ 0.25} & \tabbest{2.36 $\pm$ 0.14} & \tabbest{51.43 $\pm$ 0.57} \\
 &  & Saved mix (audit) & 3 & 44.53 $\pm$ 5.92 & 33.58 $\pm$ 16.01 & 87.15 $\pm$ 0.36 & 2.57 $\pm$ 0.12 & 46.01 $\pm$ 6.23 \\
\addlinespace[2pt]
\multirow{11}{*}{CIFAR-100} & \multirow{11}{*}{500} & ETF reference & 3 & 48.65 $\pm$ 0.68 & 23.16 $\pm$ 1.51 & 85.54 $\pm$ 0.23 & 4.29 $\pm$ 0.31 & 50.14 $\pm$ 0.51 \\
 &  & CE head (audit) & 3 & 28.56 $\pm$ 0.42 & 66.87 $\pm$ 0.55 & 86.27 $\pm$ 0.25 & 3.85 $\pm$ 0.11 & 29.39 $\pm$ 0.46 \\
 &  & Transported NCM & 3 & 50.95 $\pm$ 0.31 & 21.25 $\pm$ 0.48 & 86.94 $\pm$ 0.19 & 2.63 $\pm$ 0.19 & 52.83 $\pm$ 0.22 \\
 &  & Radial-only & 3 & 51.96 $\pm$ 0.06 & 13.84 $\pm$ 0.18 & 86.94 $\pm$ 0.19 & 2.63 $\pm$ 0.19 & 53.94 $\pm$ 0.07 \\
 &  & CompMix & 3 & 53.15 $\pm$ 0.49 & 12.71 $\pm$ 0.26 & 87.55 $\pm$ 0.15 & 2.30 $\pm$ 0.05 & 54.95 $\pm$ 0.34 \\
 &  & CompMix-Fixed & 3 & 53.07 $\pm$ 0.41 & 11.44 $\pm$ 0.26 & 87.51 $\pm$ 0.15 & 2.19 $\pm$ 0.11 & 54.86 $\pm$ 0.24 \\
 &  & InvGradMix & 3 & \tabbest{53.29 $\pm$ 0.40} & 12.76 $\pm$ 0.30 & 87.58 $\pm$ 0.07 & 2.30 $\pm$ 0.09 & \tabbest{55.11 $\pm$ 0.25} \\
 &  & InvGradMix-Fixed & 3 & 53.24 $\pm$ 0.38 & 11.42 $\pm$ 0.24 & 87.59 $\pm$ 0.07 & 2.17 $\pm$ 0.13 & \tabsecond{55.07 $\pm$ 0.20} \\
 &  & NC-ShrinkMix & 3 & 53.29 $\pm$ 0.49 & \tabsecond{11.32 $\pm$ 0.26} & \tabbest{87.76 $\pm$ 0.07} & 2.14 $\pm$ 0.12 & 55.04 $\pm$ 0.32 \\
 &  & RCR & 3 & \tabsecond{53.27 $\pm$ 0.41} & \tabbest{10.18 $\pm$ 0.38} & \tabsecond{87.74 $\pm$ 0.07} & \tabsecond{2.02 $\pm$ 0.16} & 55.04 $\pm$ 0.26 \\
 &  & Saved mix (audit) & 3 & 50.23 $\pm$ 4.17 & 24.12 $\pm$ 14.59 & 87.71 $\pm$ 0.08 & \tabbest{2.00 $\pm$ 0.25} & 51.88 $\pm$ 4.37 \\
\addlinespace[2pt]
\multirow{11}{*}{TinyImageNet} & \multirow{11}{*}{200} & ETF reference & 3 & 25.77 $\pm$ 0.50 & 43.45 $\pm$ 0.13 & 68.30 $\pm$ 0.76 & 9.58 $\pm$ 0.42 & 28.58 $\pm$ 0.38 \\
 &  & CE head (audit) & 3 & 9.91 $\pm$ 0.09 & 75.27 $\pm$ 0.33 & \tabbest{71.98 $\pm$ 0.35} & \tabbest{6.52 $\pm$ 0.09} & 12.20 $\pm$ 0.10 \\
 &  & Transported NCM & 3 & 28.71 $\pm$ 0.40 & 33.37 $\pm$ 0.45 & 68.69 $\pm$ 0.38 & 9.65 $\pm$ 0.40 & 32.17 $\pm$ 0.45 \\
 &  & Radial-only & 3 & 30.46 $\pm$ 0.20 & \tabsecond{26.66 $\pm$ 0.72} & 68.69 $\pm$ 0.38 & 9.65 $\pm$ 0.40 & 33.99 $\pm$ 0.16 \\
 &  & CompMix & 3 & 30.68 $\pm$ 0.78 & 30.89 $\pm$ 0.88 & 69.79 $\pm$ 0.99 & 8.35 $\pm$ 1.02 & 33.96 $\pm$ 0.68 \\
 &  & CompMix-Fixed & 3 & 30.18 $\pm$ 0.92 & 26.84 $\pm$ 1.29 & 68.54 $\pm$ 0.98 & 9.62 $\pm$ 1.03 & 33.74 $\pm$ 0.90 \\
 &  & InvGradMix & 3 & 30.36 $\pm$ 0.89 & 31.20 $\pm$ 0.95 & 69.75 $\pm$ 0.90 & 8.42 $\pm$ 0.93 & 33.59 $\pm$ 0.87 \\
 &  & InvGradMix-Fixed & 3 & 29.92 $\pm$ 0.91 & 27.41 $\pm$ 1.12 & 68.53 $\pm$ 0.95 & 9.76 $\pm$ 1.04 & 33.38 $\pm$ 0.86 \\
 &  & NC-ShrinkMix & 3 & \tabbest{31.55 $\pm$ 0.72} & 28.41 $\pm$ 0.83 & \tabsecond{70.55 $\pm$ 0.94} & \tabsecond{7.57 $\pm$ 0.88} & \tabsecond{34.70 $\pm$ 0.55} \\
 &  & RCR & 3 & \tabsecond{31.42 $\pm$ 0.80} & \tabbest{23.84 $\pm$ 1.15} & 69.59 $\pm$ 0.99 & 8.61 $\pm$ 1.09 & \tabbest{34.76 $\pm$ 0.68} \\
 &  & Saved mix (audit) & 3 & 31.24 $\pm$ 0.44 & 27.95 $\pm$ 0.54 & 70.23 $\pm$ 0.51 & 8.07 $\pm$ 0.48 & 34.53 $\pm$ 0.44 \\
\addlinespace[2pt]
\multirow{11}{*}{TinyImageNet} & \multirow{11}{*}{500} & ETF reference & 3 & 26.61 $\pm$ 0.39 & 41.16 $\pm$ 0.23 & 67.71 $\pm$ 0.47 & 9.78 $\pm$ 0.52 & 29.37 $\pm$ 0.52 \\
 &  & CE head (audit) & 3 & 10.81 $\pm$ 0.33 & 73.80 $\pm$ 0.38 & \tabbest{71.59 $\pm$ 0.44} & \tabbest{6.69 $\pm$ 0.60} & 13.22 $\pm$ 0.28 \\
 &  & Transported NCM & 3 & 29.59 $\pm$ 0.46 & 32.10 $\pm$ 0.40 & 69.37 $\pm$ 0.43 & 8.71 $\pm$ 0.40 & 32.92 $\pm$ 0.29 \\
 &  & Radial-only & 3 & 31.23 $\pm$ 0.08 & 25.38 $\pm$ 0.24 & 69.37 $\pm$ 0.43 & 8.71 $\pm$ 0.40 & 34.67 $\pm$ 0.02 \\
 &  & CompMix & 3 & 31.92 $\pm$ 0.01 & 28.72 $\pm$ 0.35 & 70.32 $\pm$ 0.32 & 7.71 $\pm$ 0.37 & 35.16 $\pm$ 0.15 \\
 &  & CompMix-Fixed & 3 & 31.34 $\pm$ 0.15 & \tabsecond{25.20 $\pm$ 0.25} & 69.21 $\pm$ 0.20 & 8.91 $\pm$ 0.20 & 34.71 $\pm$ 0.19 \\
 &  & InvGradMix & 3 & 31.68 $\pm$ 0.12 & 29.10 $\pm$ 0.53 & 70.47 $\pm$ 0.41 & 7.68 $\pm$ 0.50 & 34.91 $\pm$ 0.05 \\
 &  & InvGradMix-Fixed & 3 & 31.09 $\pm$ 0.21 & 25.78 $\pm$ 0.18 & 69.32 $\pm$ 0.18 & 8.87 $\pm$ 0.17 & 34.44 $\pm$ 0.11 \\
 &  & NC-ShrinkMix & 3 & \tabbest{32.47 $\pm$ 0.17} & 26.67 $\pm$ 0.23 & \tabsecond{70.89 $\pm$ 0.44} & \tabsecond{7.10 $\pm$ 0.56} & \tabbest{35.63 $\pm$ 0.27} \\
 &  & RCR & 3 & \tabsecond{32.24 $\pm$ 0.05} & \tabbest{22.47 $\pm$ 0.13} & 70.16 $\pm$ 0.20 & 7.90 $\pm$ 0.16 & \tabsecond{35.46 $\pm$ 0.08} \\
 &  & Saved mix (audit) & 3 & 31.93 $\pm$ 0.13 & 26.80 $\pm$ 0.36 & 70.53 $\pm$ 0.42 & 7.48 $\pm$ 0.40 & 35.29 $\pm$ 0.12 \\
\bottomrule
\end{tabular}%
}
\renewcommand{\arraystretch}{1.0}
\tabnote{Bold marks the best mean and underline the second best mean within each dataset/buffer block. Saved mix is included as an audit row because it depends on the score stored at artifact generation time.}
\end{table}

%% file: tables_tex/nct_diagnostic.tex
\begin{table}[t]
\centering
\caption{\textbf{Diagnostic check on a second progressive-ETF learner.}
We reproduce an NCT-style progressive ETF run on Seq-CIFAR-100 with buffer
200 and evaluate an offline RCR-style ramp correction on the same saved
features. The ramp keeps ETF evidence dominant at early tasks and increases
the radial correction later. The purpose is diagnostic rather than a full
re-benchmarking of NCT: it checks that the route-correction signal is not
unique to the ProNC implementation.}
\label{tab:nct_diagnostic}
\scriptsize
\setlength{\tabcolsep}{4pt}
\begin{tabular}{@{}lccccc@{}}
\toprule
Head & Class FAA $\uparrow$ & Class FF $\downarrow$ & Task FAA $\uparrow$ & Task FF $\downarrow$ & Route $\uparrow$ \\
\midrule
NCT-style ETF & 31.62 & 27.05 & 79.93 & 4.99 & 33.58 \\
NCT-style ETF + RCR-style ramp & 34.02 & 25.84 & 81.33 & 3.31 & 35.89 \\
\midrule
$\Delta$ & $+2.40$ & $-1.21$ & $+1.41$ & $-1.68$ & $+2.32$ \\
\bottomrule
\end{tabular}
\end{table}

%% file: tables_tex/score_component_ablation.tex
\begin{table}[H]
  \centering
  \caption{\textbf{Transport, radial reliability, and NC shrinkage are complementary.}}
  \label{tab:score_component_ablation}
  \scriptsize
  \setlength{\tabcolsep}{3.0pt}
  \begin{tabular}{llrrr}
    \toprule
    Dataset/Buffer & Head & Class-IL $\uparrow$ & Task-IL $\uparrow$ & Route $\uparrow$ \\
    \midrule
    Seq-CIFAR-10 / 200 & ETF only & $66.14\pm1.38$ & $96.74\pm0.10$ & $66.58\pm1.34$ \\
     & Transported NCM & $63.46\pm0.61$ & $96.11\pm0.43$ & $63.98\pm0.54$ \\
     & Radial only & $64.06\pm0.42$ & $96.27\pm0.15$ & $64.66\pm0.32$ \\
     & ETF+NCM & $66.31\pm0.82$ & $96.55\pm0.26$ & $66.80\pm0.70$ \\
     & ETF+raw radial & $68.50\pm0.33$ & $96.62\pm0.08$ & $69.05\pm0.29$ \\
     & ETF+equal radius & $66.31\pm0.82$ & $96.55\pm0.26$ & $66.80\pm0.70$ \\
     & ETF+no transport & $66.14\pm1.38$ & $96.76\pm0.09$ & $66.58\pm1.34$ \\
    \oursrow
     & \textbf{Ours (\method)} & $70.36\pm0.50$ & $96.67\pm0.08$ & $70.82\pm0.46$ \\
    \addlinespace[1pt]
    Seq-CIFAR-10 / 500 & ETF only & $72.16\pm1.30$ & $96.82\pm0.12$ & $72.71\pm1.35$ \\
     & Transported NCM & $66.28\pm0.27$ & $96.34\pm0.03$ & $66.95\pm0.29$ \\
     & Radial only & $68.90\pm0.81$ & $96.24\pm0.02$ & $69.65\pm0.69$ \\
     & ETF+NCM & $71.73\pm0.69$ & $96.72\pm0.04$ & $72.32\pm0.61$ \\
     & ETF+raw radial & $71.90\pm1.64$ & $96.73\pm0.13$ & $72.61\pm1.65$ \\
     & ETF+equal radius & $71.73\pm0.69$ & $96.72\pm0.04$ & $72.32\pm0.61$ \\
     & ETF+no transport & $72.16\pm1.30$ & $96.83\pm0.11$ & $72.71\pm1.35$ \\
    \oursrow
     & \textbf{Ours (\method)} & $73.68\pm0.72$ & $96.81\pm0.03$ & $74.33\pm0.83$ \\
    \addlinespace[1pt]
    Seq-CIFAR-100 / 200 & ETF only & $41.71\pm0.61$ & $85.53\pm0.16$ & $42.81\pm0.57$ \\
     & Transported NCM & $43.59\pm0.65$ & $85.64\pm0.07$ & $45.41\pm0.64$ \\
     & Radial only & $46.21\pm0.44$ & $85.46\pm0.26$ & $48.36\pm0.49$ \\
     & ETF+NCM & $44.49\pm0.29$ & $86.64\pm0.18$ & $46.04\pm0.28$ \\
     & ETF+raw radial & $43.10\pm1.07$ & $86.57\pm0.42$ & $44.87\pm1.05$ \\
     & ETF+equal radius & $44.49\pm0.29$ & $86.64\pm0.18$ & $46.04\pm0.28$ \\
     & ETF+no transport & $41.71\pm0.61$ & $85.56\pm0.16$ & $42.81\pm0.57$ \\
    \oursrow
     & \textbf{Ours (\method)} & $49.74\pm0.58$ & $87.28\pm0.20$ & $51.43\pm0.47$ \\
    \addlinespace[1pt]
    Seq-CIFAR-100 / 500 & ETF only & $48.65\pm0.55$ & $85.54\pm0.19$ & $50.14\pm0.42$ \\
     & Transported NCM & $50.95\pm0.25$ & $86.94\pm0.15$ & $52.83\pm0.18$ \\
     & Radial only & $50.93\pm0.22$ & $86.65\pm0.10$ & $52.88\pm0.19$ \\
     & ETF+NCM & $51.47\pm0.18$ & $87.48\pm0.12$ & $53.26\pm0.10$ \\
     & ETF+raw radial & $45.98\pm1.03$ & $87.09\pm0.11$ & $47.66\pm0.96$ \\
     & ETF+equal radius & $51.47\pm0.18$ & $87.48\pm0.12$ & $53.26\pm0.10$ \\
     & ETF+no transport & $48.65\pm0.55$ & $85.61\pm0.18$ & $50.14\pm0.42$ \\
    \oursrow
     & \textbf{Ours (\method)} & $53.27\pm0.34$ & $87.74\pm0.05$ & $55.04\pm0.21$ \\
    \addlinespace[1pt]
    Seq-TinyImageNet / 200 & ETF only & $25.77\pm0.41$ & $68.30\pm0.62$ & $28.58\pm0.31$ \\
     & Transported NCM & $28.71\pm0.33$ & $68.69\pm0.31$ & $32.17\pm0.37$ \\
     & Radial only & $28.21\pm0.42$ & $66.09\pm1.00$ & $32.14\pm0.33$ \\
     & ETF+NCM & $29.29\pm0.41$ & $70.01\pm0.35$ & $32.63\pm0.37$ \\
     & ETF+raw radial & $19.96\pm2.18$ & $54.13\pm2.77$ & $24.79\pm1.02$ \\
     & ETF+equal radius & $29.29\pm0.41$ & $70.01\pm0.35$ & $32.63\pm0.37$ \\
     & ETF+no transport & $25.77\pm0.41$ & $68.31\pm0.68$ & $28.58\pm0.31$ \\
    \oursrow
     & \textbf{Ours (\method)} & $31.42\pm0.65$ & $69.59\pm0.81$ & $34.76\pm0.56$ \\
    \addlinespace[1pt]
    Seq-TinyImageNet / 500 & ETF only & $26.61\pm0.32$ & $67.71\pm0.39$ & $29.37\pm0.42$ \\
     & Transported NCM & $29.59\pm0.37$ & $69.37\pm0.35$ & $32.92\pm0.24$ \\
     & Radial only & $30.00\pm0.19$ & $67.33\pm0.13$ & $33.72\pm0.24$ \\
     & ETF+NCM & $29.95\pm0.28$ & $70.34\pm0.30$ & $33.27\pm0.17$ \\
     & ETF+raw radial & $24.14\pm0.90$ & $58.26\pm0.53$ & $28.37\pm0.66$ \\
     & ETF+equal radius & $29.95\pm0.28$ & $70.34\pm0.30$ & $33.27\pm0.17$ \\
     & ETF+no transport & $26.62\pm0.33$ & $67.72\pm0.41$ & $29.38\pm0.43$ \\
    \oursrow
     & \textbf{Ours (\method)} & $32.24\pm0.04$ & $70.16\pm0.16$ & $35.46\pm0.07$ \\
    \addlinespace[1pt]
    \bottomrule
  \end{tabular}
  \tabnote{All heads use the same frozen features. No-transport uses the ETF vertex as the radial center, isolating the role of transported class measures.}
\end{table}

%% file: tables_tex/omega_sensitivity.tex
\begin{table}[H]
  \centering
  \caption{\textbf{Sensitivity to the ETF/NC-shrunk radial mixture weight.}}
  \label{tab:omega_sensitivity}
  \scriptsize
  \setlength{\tabcolsep}{2.2pt}
  \begin{minipage}[t]{0.49\linewidth}
  \centering
  \resizebox{\linewidth}{!}{%
  \begin{tabular}{llrrr}
    \toprule
    Set & $\omega$ & Class & Task & Route \\
    \midrule
    C10/200 & 0.00 & $66.14\pm1.38$ & $96.74\pm0.10$ & $66.58\pm1.34$ \\
     & 0.25 & $67.49\pm1.15$ & $96.82\pm0.09$ & $67.93\pm1.11$ \\
     & 0.50 & $69.09\pm0.70$ & $96.78\pm0.11$ & $69.56\pm0.62$ \\
     & 0.75 & $70.72\pm0.56$ & $96.62\pm0.08$ & $71.19\pm0.53$ \\
    \oursrow
     & \textbf{Ours} & $70.36\pm0.50$ & $96.67\pm0.08$ & $70.82\pm0.46$ \\
     & 1.00 & $67.09\pm0.34$ & $96.43\pm0.14$ & $67.62\pm0.24$ \\
    \addlinespace[1pt]
    C10/500 & 0.00 & $72.16\pm1.30$ & $96.82\pm0.12$ & $72.71\pm1.35$ \\
     & 0.25 & $73.53\pm1.45$ & $96.88\pm0.10$ & $74.09\pm1.49$ \\
     & 0.50 & $73.81\pm1.02$ & $96.88\pm0.04$ & $74.42\pm1.08$ \\
     & 0.75 & $73.48\pm0.51$ & $96.75\pm0.02$ & $74.15\pm0.63$ \\
    \oursrow
     & \textbf{Ours} & $73.68\pm0.72$ & $96.81\pm0.03$ & $74.33\pm0.83$ \\
     & 1.00 & $70.82\pm0.39$ & $96.41\pm0.01$ & $71.52\pm0.25$ \\
    \addlinespace[1pt]
    C100/200 & 0.00 & $41.71\pm0.61$ & $85.53\pm0.16$ & $42.81\pm0.57$ \\
     & 0.25 & $43.48\pm0.42$ & $86.34\pm0.20$ & $44.63\pm0.39$ \\
     & 0.50 & $45.76\pm0.28$ & $87.10\pm0.15$ & $47.01\pm0.14$ \\
     & 0.75 & $48.33\pm0.52$ & $87.73\pm0.22$ & $49.74\pm0.40$ \\
    \oursrow
     & \textbf{Ours} & $49.74\pm0.58$ & $87.28\pm0.20$ & $51.43\pm0.47$ \\
     & 1.00 & $48.08\pm0.48$ & $86.34\pm0.20$ & $49.98\pm0.44$ \\
    \bottomrule
  \end{tabular}%
  }
  \end{minipage}\hfill
  \begin{minipage}[t]{0.49\linewidth}
  \centering
  \resizebox{\linewidth}{!}{%
  \begin{tabular}{llrrr}
    \toprule
    Set & $\omega$ & Class & Task & Route \\
    \midrule
    C100/500 & 0.00 & $48.65\pm0.55$ & $85.54\pm0.19$ & $50.14\pm0.42$ \\
     & 0.25 & $49.90\pm0.51$ & $86.48\pm0.15$ & $51.46\pm0.38$ \\
     & 0.50 & $51.19\pm0.25$ & $87.35\pm0.10$ & $52.76\pm0.11$ \\
     & 0.75 & $52.86\pm0.35$ & $87.90\pm0.07$ & $54.53\pm0.19$ \\
    \oursrow
     & \textbf{Ours} & $53.27\pm0.34$ & $87.74\pm0.05$ & $55.04\pm0.21$ \\
     & 1.00 & $51.72\pm0.38$ & $87.10\pm0.08$ & $53.58\pm0.28$ \\
    \addlinespace[1pt]
    Tiny/200 & 0.00 & $25.77\pm0.41$ & $68.30\pm0.62$ & $28.58\pm0.31$ \\
     & 0.25 & $26.82\pm0.50$ & $69.70\pm0.63$ & $29.65\pm0.48$ \\
     & 0.50 & $28.16\pm0.41$ & $70.95\pm0.41$ & $31.15\pm0.33$ \\
     & 0.75 & $30.28\pm0.59$ & $71.38\pm0.50$ & $33.29\pm0.48$ \\
    \oursrow
     & \textbf{Ours} & $31.42\pm0.65$ & $69.59\pm0.81$ & $34.76\pm0.56$ \\
     & 1.00 & $30.38\pm0.47$ & $68.12\pm0.72$ & $34.03\pm0.51$ \\
    \addlinespace[1pt]
    Tiny/500 & 0.00 & $26.61\pm0.32$ & $67.71\pm0.39$ & $29.37\pm0.42$ \\
     & 0.25 & $27.80\pm0.29$ & $69.21\pm0.51$ & $30.62\pm0.40$ \\
     & 0.50 & $29.23\pm0.30$ & $70.64\pm0.50$ & $32.12\pm0.46$ \\
     & 0.75 & $31.31\pm0.25$ & $71.50\pm0.32$ & $34.30\pm0.38$ \\
    \oursrow
     & \textbf{Ours} & $32.24\pm0.04$ & $70.16\pm0.16$ & $35.46\pm0.07$ \\
     & 1.00 & $31.44\pm0.25$ & $68.92\pm0.12$ & $34.90\pm0.17$ \\
    \bottomrule
  \end{tabular}%
  }
  \end{minipage}
  \tabnote{Class, Task, and Route report final accuracies. The public method uses the deterministic fixed value $\omega=q/(q+1)$, not a value selected on test data.}
\end{table}

%% file: tables_tex/rcr_schedule_ablation.tex
\begin{table}[H]
  \centering
  \caption{\textbf{Ablation of the RCR mixture schedule on Seq-CIFAR-100.}
  The deployed fixed-cap rule uses $\omega=q/(q+1)$ at every task, while the stage schedule uses $\omega_t=\min(q/(q+1),(t-1)/t)$. Both variants use the NC-shrunk radial evidence and are evaluated on the same frozen artifacts; lower FF is better.}
  \label{tab:rcr_schedule_ablation}
  \footnotesize
  \begin{tabular}{llrrrrr}
    \toprule
    Buffer & Mixture rule & $n$ & Class FAA & Class FF & Task FAA & Route \\
    \midrule
    200 & Stage schedule & 2 & 49.55 & 19.31 & 87.49 & 51.21 \\
    \oursrow
    200 & \textbf{Ours: fixed cap} & 2 & 49.49 & 16.43 & 87.39 & 51.18 \\
    \addlinespace[1pt]
    500 & Stage schedule & 2 & 53.12 & 11.44 & 87.72 & 54.91 \\
    \oursrow
    500 & \textbf{Ours: fixed cap} & 2 & 53.13 & 10.29 & 87.71 & 54.94 \\
    \bottomrule
  \end{tabular}
  \tabnote{The fixed-cap rule removes the deterministic stage ramp and is used by the reported Ours head. It reaches nearly the same final routing as the staged variant on these runs, while applying maximum radial evidence from the first task.}
\end{table}

%% file: tables_tex/alpha_rho_sensitivity.tex
\begin{table}[H]
  \centering
  \caption{\textbf{Sensitivity to the competitive-radius exponent.}
  Seq-CIFAR-100 buffer 200, final task. The default $\alpha_\rho=0.25$ is stable: larger changes in the exponent preserve the routing gain, while the exact value is not selected on the test set.}
  \label{tab:alpha_rho_sensitivity}
  \scriptsize
  \setlength{\tabcolsep}{4.2pt}
  \begin{tabular}{lrrrr}
    \toprule
    $\alpha_\rho$ & Class-IL $\uparrow$ & Task-IL $\uparrow$ & Route $\uparrow$ & Wrong-task $\downarrow$ \\
    \midrule
    0.00 & $46.44\pm0.46$ & $87.06\pm0.22$ & $48.02\pm0.45$ & $51.98\pm0.45$ \\
    0.10 & $48.56\pm0.43$ & $87.15\pm0.17$ & $50.18\pm0.35$ & $49.82\pm0.35$ \\
    \oursrow
    \textbf{Ours: 0.25} & $49.74\pm0.58$ & $87.28\pm0.20$ & $51.43\pm0.47$ & $48.57\pm0.47$ \\
    0.50 & $48.54\pm0.78$ & $87.35\pm0.21$ & $50.27\pm0.70$ & $49.73\pm0.70$ \\
    1.00 & $43.61\pm1.05$ & $86.76\pm0.36$ & $45.34\pm1.01$ & $54.66\pm1.01$ \\
    \bottomrule
  \end{tabular}
\end{table}

%% file: tables_tex/shrink_clip_ablation.tex
\begin{table}[H]
  \centering
  \caption{\textbf{Upper clipping for NC retraction.}
  We set $\alpha_{\mathrm{lo}}=0$ and compare the deployed upper safeguard
  $\alpha_{\mathrm{hi}}=0.95$ to the unconstrained endpoint
  $\alpha_{\mathrm{hi}}=1$. The lower clip is inactive on these runs; the
  useful safeguard is preventing full trust in transported centers.}
  \label{tab:shrink_clip_ablation}
  \scriptsize
  \setlength{\tabcolsep}{4.0pt}
  \begin{tabular}{llrrr}
    \toprule
    Dataset / Buffer & Setting & $\alpha_{\mathrm{hi}}$ &
    Class FAA $\uparrow$ & Class FF $\downarrow$ \\
    \midrule
    Seq-CIFAR-10 / 200 & No upper safeguard & 1.00 & $70.44\pm0.59$ & $22.63\pm1.14$ \\
    \oursrow
    Seq-CIFAR-10 / 200 & \textbf{Ours} & 0.95 & $70.36\pm0.50$ & $22.85\pm1.19$ \\
    \addlinespace[1pt]
    Seq-CIFAR-10 / 500 & No upper safeguard & 1.00 & $73.56\pm0.60$ & $17.60\pm0.72$ \\
    \oursrow
    Seq-CIFAR-10 / 500 & \textbf{Ours} & 0.95 & $73.68\pm0.72$ & $17.73\pm0.72$ \\
    \addlinespace[1pt]
    Seq-CIFAR-100 / 200 & No upper safeguard & 1.00 & $49.41\pm0.58$ & $17.46\pm0.51$ \\
    \oursrow
    Seq-CIFAR-100 / 200 & \textbf{Ours} & 0.95 & $49.74\pm0.58$ & $16.27\pm0.37$ \\
    \addlinespace[1pt]
    Seq-CIFAR-100 / 500 & No upper safeguard & 1.00 & $53.06\pm0.40$ & $10.96\pm0.28$ \\
    \oursrow
    Seq-CIFAR-100 / 500 & \textbf{Ours} & 0.95 & $53.27\pm0.34$ & $10.18\pm0.31$ \\
    \addlinespace[1pt]
    Seq-TinyImageNet / 200 & No upper safeguard & 1.00 & $30.57\pm0.58$ & $26.46\pm0.82$ \\
    \oursrow
    Seq-TinyImageNet / 200 & \textbf{Ours} & 0.95 & $31.42\pm0.65$ & $23.84\pm0.94$ \\
    \addlinespace[1pt]
    Seq-TinyImageNet / 500 & No upper safeguard & 1.00 & $31.45\pm0.12$ & $25.12\pm0.20$ \\
    \oursrow
    Seq-TinyImageNet / 500 & \textbf{Ours} & 0.95 & $32.24\pm0.04$ & $22.47\pm0.10$ \\
    \bottomrule
  \end{tabular}
\end{table}

%% file: tables_tex/calibration_baselines_all.tex
\begin{table}[t]
  \centering
  \caption{\textbf{Replay-fitted calibration does not explain RCR.}}
  \label{tab:calibration_baselines_all}
  \scriptsize
  \setlength{\tabcolsep}{3.0pt}
  \begin{tabular}{llrrrr}
    \toprule
    Dataset/Buffer & Head & Class-IL $\uparrow$ & Task-IL $\uparrow$ & Route $\uparrow$ & Wrong-task $\downarrow$ \\
    \midrule
    Seq-CIFAR-10 / 200 & ProNC ETF & $66.14\pm1.38$ & $96.74\pm0.10$ & $66.58\pm1.34$ & $33.42\pm1.34$ \\
     & Replay task-bias & $59.21\pm0.65$ & $96.74\pm0.10$ & $59.60\pm0.64$ & $40.40\pm0.64$ \\
     & Replay class-bias & $59.36\pm0.62$ & $96.72\pm0.12$ & $59.75\pm0.60$ & $40.25\pm0.60$ \\
     & Replay task affine & $58.19\pm1.35$ & $96.74\pm0.10$ & $58.61\pm1.32$ & $41.39\pm1.32$ \\
    \oursrow
     & \textbf{Ours (\method)} & $70.36\pm0.50$ & $96.67\pm0.08$ & $70.82\pm0.46$ & $29.18\pm0.46$ \\
    \addlinespace[1pt]
    Seq-CIFAR-10 / 500 & ProNC ETF & $72.16\pm1.30$ & $96.82\pm0.12$ & $72.71\pm1.35$ & $27.29\pm1.35$ \\
     & Replay task-bias & $67.89\pm0.91$ & $96.82\pm0.12$ & $68.35\pm0.97$ & $31.65\pm0.97$ \\
     & Replay class-bias & $68.03\pm0.88$ & $96.87\pm0.07$ & $68.49\pm0.93$ & $31.51\pm0.93$ \\
     & Replay task affine & $67.84\pm0.49$ & $96.82\pm0.12$ & $68.33\pm0.47$ & $31.67\pm0.47$ \\
    \oursrow
     & \textbf{Ours (\method)} & $73.68\pm0.72$ & $96.81\pm0.03$ & $74.33\pm0.83$ & $25.67\pm0.83$ \\
    \addlinespace[1pt]
    Seq-CIFAR-100 / 200 & ProNC ETF & $41.71\pm0.61$ & $85.53\pm0.16$ & $42.81\pm0.57$ & $57.19\pm0.57$ \\
     & Replay task-bias & $29.91\pm0.63$ & $85.53\pm0.16$ & $30.67\pm0.59$ & $69.33\pm0.59$ \\
     & Replay class-bias & $29.36\pm0.65$ & $84.86\pm0.06$ & $30.13\pm0.70$ & $69.87\pm0.70$ \\
     & Replay task affine & $27.54\pm2.57$ & $85.53\pm0.16$ & $28.35\pm2.56$ & $71.65\pm2.56$ \\
    \oursrow
     & \textbf{Ours (\method)} & $49.74\pm0.58$ & $87.28\pm0.20$ & $51.43\pm0.47$ & $48.57\pm0.47$ \\
    \addlinespace[1pt]
    Seq-CIFAR-100 / 500 & ProNC ETF & $48.65\pm0.55$ & $85.54\pm0.19$ & $50.14\pm0.42$ & $49.86\pm0.42$ \\
     & Replay task-bias & $36.14\pm0.85$ & $85.54\pm0.19$ & $37.06\pm0.91$ & $62.94\pm0.91$ \\
     & Replay class-bias & $36.22\pm0.35$ & $85.51\pm0.36$ & $37.12\pm0.29$ & $62.88\pm0.29$ \\
     & Replay task affine & $36.03\pm0.89$ & $85.54\pm0.19$ & $37.05\pm0.82$ & $62.95\pm0.82$ \\
    \oursrow
     & \textbf{Ours (\method)} & $53.27\pm0.34$ & $87.74\pm0.05$ & $55.04\pm0.21$ & $44.96\pm0.21$ \\
    \addlinespace[1pt]
    Seq-TinyImageNet / 200 & ProNC ETF & $25.77\pm0.41$ & $68.30\pm0.62$ & $28.58\pm0.31$ & $71.42\pm0.31$ \\
     & Replay task-bias & $28.48\pm0.11$ & $68.30\pm0.62$ & $31.43\pm0.11$ & $68.57\pm0.11$ \\
     & Replay class-bias & $29.49\pm0.15$ & $72.05\pm0.21$ & $32.50\pm0.14$ & $67.50\pm0.14$ \\
     & Replay task affine & $31.42\pm0.45$ & $68.30\pm0.62$ & $34.48\pm0.40$ & $65.52\pm0.40$ \\
    \oursrow
     & \textbf{Ours (\method)} & $31.42\pm0.65$ & $69.59\pm0.81$ & $34.76\pm0.56$ & $65.24\pm0.56$ \\
    \addlinespace[1pt]
    Seq-TinyImageNet / 500 & ProNC ETF & $26.61\pm0.32$ & $67.71\pm0.39$ & $29.37\pm0.42$ & $70.63\pm0.42$ \\
     & Replay task-bias & $29.80\pm0.16$ & $67.71\pm0.39$ & $32.59\pm0.17$ & $67.41\pm0.17$ \\
     & Replay class-bias & $30.37\pm0.32$ & $71.54\pm0.25$ & $33.28\pm0.27$ & $66.72\pm0.27$ \\
     & Replay task affine & $32.02\pm0.28$ & $67.71\pm0.39$ & $35.14\pm0.20$ & $64.86\pm0.20$ \\
    \oursrow
     & \textbf{Ours (\method)} & $32.24\pm0.04$ & $70.16\pm0.16$ & $35.46\pm0.07$ & $64.54\pm0.07$ \\
    \addlinespace[1pt]
    \bottomrule
  \end{tabular}
  \tabnote{All calibration heads are fit only on replay-buffer scores and labels, then evaluated on held-out test features.}
\end{table}

%% file: tables_tex/nc_deviation_diag.tex
\begin{table}[t]
  \centering
  \caption{\textbf{Neural-collapse deviations and routing drift on Seq-CIFAR-100, buffer 200.}
  Panel (a) tracks stage-level NC diagnostics and held-out performance. Panel (b) reports final-stage class-wise Pearson correlations: prototype--ETF deviation and competitive radius identify older classes with larger route error and larger \method gains, while NC1 dispersion mostly tracks age. Values are mean $\pm$ std over saved runs.}
  \label{tab:nc_deviation_diag}
  \scriptsize
  \setlength{\tabcolsep}{2.6pt}
  {\small\textcolor{trolleygrey}{\textit{(a) Stage-level diagnostics}}}\par
  \begin{tabular}{lrrrrr}
    \toprule
    Stage & $\langle p,u\rangle\uparrow$ & NC1 disp. $\downarrow$ & CV$(\rho)$ & Route $\uparrow$ & Class-IL $\uparrow$ \\
    \midrule
    T3 & $0.39\pm0.00$ & $0.27\pm0.00$ & $0.08\pm0.01$ & $75.34\pm1.01$ $\to$ $74.79\pm0.83$ & $70.98\pm0.89$ $\to$ $70.58\pm0.63$ \\
    T10 & $0.19\pm0.00$ & $0.41\pm0.00$ & $0.11\pm0.00$ & $42.81\pm0.57$ $\to$ $51.43\pm0.47$ & $41.71\pm0.61$ $\to$ $49.74\pm0.58$ \\
    \bottomrule
  \end{tabular} \\
  {\small\textcolor{trolleygrey}{\textit{(b) Final-stage class-wise correlations}}}\par
  \begin{tabular}{lrrr}
    \toprule
    Class signal & Route error & $\Delta$Class & Age \\
    \midrule
    $\delta_c=d_{\cos}(p_c,u_c)$ & $0.68\pm0.02$ & $0.59\pm0.03$ & $0.90\pm0.02$ \\
    $\rho_c$ & $0.61\pm0.00$ & $0.55\pm0.05$ & $0.94\pm0.01$ \\
    NC1 disp. & $0.24\pm0.03$ & $0.13\pm0.04$ & $0.81\pm0.01$ \\
    \bottomrule
  \end{tabular}
  \tabnote{NC1 dispersion is the held-out mean $d_{\cos}(z,p_c)$ within each class. Route error and $\Delta$Class are computed per class before correlation.}
\end{table}

%% file: tables_tex/route_oracle_bounds.tex
\begin{table}[t]
  \centering
  \caption{\textbf{Route-oracle bounds.}}
  \label{tab:route_oracle_bounds}
  \scriptsize
  \setlength{\tabcolsep}{3.0pt}
  \begin{tabular}{llrrr}
    \toprule
    Dataset/Buffer & Head & Class-IL $\uparrow$ & Task-IL $\uparrow$ & Route $\uparrow$ \\
    \midrule
    Seq-CIFAR-10 / 200 & ProNC ETF & $66.14\pm1.38$ & $96.74\pm0.10$ & $66.58\pm1.34$ \\
     & RCR & $70.36\pm0.50$ & $96.67\pm0.08$ & $70.82\pm0.46$ \\
     & ProNC + true route & $96.74\pm0.10$ & $96.74\pm0.10$ & $100.00\pm0.00$ \\
     & RCR + true route & $96.67\pm0.08$ & $96.67\pm0.08$ & $100.00\pm0.00$ \\
    \addlinespace[1pt]
    Seq-CIFAR-10 / 500 & ProNC ETF & $72.16\pm1.30$ & $96.82\pm0.12$ & $72.71\pm1.35$ \\
     & RCR & $73.68\pm0.72$ & $96.81\pm0.03$ & $74.33\pm0.83$ \\
     & ProNC + true route & $96.82\pm0.12$ & $96.82\pm0.12$ & $100.00\pm0.00$ \\
     & RCR + true route & $96.81\pm0.03$ & $96.81\pm0.03$ & $100.00\pm0.00$ \\
    \addlinespace[1pt]
    Seq-CIFAR-100 / 200 & ProNC ETF & $41.71\pm0.61$ & $85.53\pm0.16$ & $42.81\pm0.57$ \\
     & RCR & $49.74\pm0.58$ & $87.28\pm0.20$ & $51.43\pm0.47$ \\
     & ProNC + true route & $85.53\pm0.16$ & $85.53\pm0.16$ & $100.00\pm0.00$ \\
     & RCR + true route & $87.28\pm0.20$ & $87.28\pm0.20$ & $100.00\pm0.00$ \\
    \addlinespace[1pt]
    Seq-CIFAR-100 / 500 & ProNC ETF & $48.65\pm0.55$ & $85.54\pm0.19$ & $50.14\pm0.42$ \\
     & RCR & $53.27\pm0.34$ & $87.74\pm0.05$ & $55.04\pm0.21$ \\
     & ProNC + true route & $85.54\pm0.19$ & $85.54\pm0.19$ & $100.00\pm0.00$ \\
     & RCR + true route & $87.74\pm0.05$ & $87.74\pm0.05$ & $100.00\pm0.00$ \\
    \addlinespace[1pt]
    Seq-TinyImageNet / 200 & ProNC ETF & $25.77\pm0.41$ & $68.30\pm0.62$ & $28.58\pm0.31$ \\
     & RCR & $31.42\pm0.65$ & $69.59\pm0.81$ & $34.76\pm0.56$ \\
     & ProNC + true route & $68.30\pm0.62$ & $68.30\pm0.62$ & $100.00\pm0.00$ \\
     & RCR + true route & $69.59\pm0.81$ & $69.59\pm0.81$ & $100.00\pm0.00$ \\
    \addlinespace[1pt]
    Seq-TinyImageNet / 500 & ProNC ETF & $26.61\pm0.32$ & $67.71\pm0.39$ & $29.37\pm0.42$ \\
     & RCR & $32.24\pm0.04$ & $70.16\pm0.16$ & $35.46\pm0.07$ \\
     & ProNC + true route & $67.71\pm0.39$ & $67.71\pm0.39$ & $100.00\pm0.00$ \\
     & RCR + true route & $70.16\pm0.16$ & $70.16\pm0.16$ & $100.00\pm0.00$ \\
    \addlinespace[1pt]
    \bottomrule
  \end{tabular}
  \tabnote{The true-route rows restrict the argmax to the ground-truth task block and are upper bounds, not deployable methods.}
\end{table}

%% file: tables_tex/classwise_nc_correlation.tex
\begin{table}[t]
  \centering
  \caption{\textbf{Class-wise NC-deviation correlations.}}
  \label{tab:classwise_nc_correlation}
  \scriptsize
  \setlength{\tabcolsep}{3.0pt}
  \begin{tabular}{llrr}
    \toprule
    Dataset/Buffer & Signal & Route error $r$ & $\Delta$Class $r$ \\
    \midrule
    Seq-CIFAR-10 / 200 & $d_{\cos}(p_c,u_c)$ & $0.76\pm0.11$ & $0.39\pm0.16$ \\
     & $\rho_c$ & $0.76\pm0.16$ & $0.36\pm0.17$ \\
     & NC1 disp. & $0.70\pm0.11$ & $0.33\pm0.11$ \\
     & Age & $0.45\pm0.26$ & $0.31\pm0.32$ \\
    \addlinespace[1pt]
    Seq-CIFAR-10 / 500 & $d_{\cos}(p_c,u_c)$ & $0.72\pm0.13$ & $0.49\pm0.17$ \\
     & $\rho_c$ & $0.73\pm0.16$ & $0.49\pm0.17$ \\
     & NC1 disp. & $0.67\pm0.15$ & $0.39\pm0.19$ \\
     & Age & $0.42\pm0.29$ & $0.54\pm0.20$ \\
    \addlinespace[1pt]
    Seq-CIFAR-100 / 200 & $d_{\cos}(p_c,u_c)$ & $0.68\pm0.02$ & $0.59\pm0.03$ \\
     & $\rho_c$ & $0.61\pm0.00$ & $0.55\pm0.05$ \\
     & NC1 disp. & $0.24\pm0.03$ & $0.13\pm0.04$ \\
     & Age & $0.46\pm0.01$ & $0.38\pm0.05$ \\
    \addlinespace[1pt]
    Seq-CIFAR-100 / 500 & $d_{\cos}(p_c,u_c)$ & $0.61\pm0.01$ & $0.62\pm0.02$ \\
     & $\rho_c$ & $0.46\pm0.02$ & $0.55\pm0.02$ \\
     & NC1 disp. & $0.21\pm0.00$ & $0.14\pm0.02$ \\
     & Age & $0.27\pm0.02$ & $0.36\pm0.02$ \\
    \addlinespace[1pt]
    Seq-TinyImageNet / 200 & $d_{\cos}(p_c,u_c)$ & $0.76\pm0.01$ & $0.41\pm0.05$ \\
     & $\rho_c$ & $0.50\pm0.03$ & $0.43\pm0.04$ \\
     & NC1 disp. & $0.44\pm0.02$ & $0.08\pm0.02$ \\
     & Age & $0.57\pm0.04$ & $0.16\pm0.07$ \\
    \addlinespace[1pt]
    Seq-TinyImageNet / 500 & $d_{\cos}(p_c,u_c)$ & $0.71\pm0.00$ & $0.44\pm0.01$ \\
     & $\rho_c$ & $0.54\pm0.03$ & $0.42\pm0.06$ \\
     & NC1 disp. & $0.47\pm0.01$ & $0.11\pm0.03$ \\
     & Age & $0.54\pm0.02$ & $0.21\pm0.02$ \\
    \addlinespace[1pt]
    \bottomrule
  \end{tabular}
  \tabnote{Pearson correlations are computed per seed across classes and then averaged.}
\end{table}